\documentclass[journal]{IEEEtran}
\ifCLASSINFOpdf
   \usepackage[pdftex]{graphicx}
\else
\fi

\usepackage[dvipsnames]{xcolor}
\usepackage{url}
\usepackage{bm}
\usepackage{multirow}
\usepackage{tabularx}
\usepackage{color}
\usepackage{hyperref}
\usepackage{caption}
\usepackage{soul}

\usepackage{booktabs}

\usepackage{times}
\usepackage{epsfig}
\usepackage{graphicx}
\usepackage{amsmath}
\usepackage{amssymb}
\usepackage{comment}
\usepackage{authblk}
\usepackage{multirow}
\usepackage{tabularx}
\usepackage{balance}
\usepackage{subfigure}
\usepackage{balance}
\usepackage{algorithm}
\usepackage{xcolor}
\usepackage{color, colortbl}
\usepackage[noend]{algpseudocode}
\usepackage{cite}
\usepackage{enumitem}
\setlist{nosep}
\usepackage{diagbox}

\usepackage{pifont}
\newcommand{\cmark}{\ding{51}}%
\newcommand{\xmark}{\ding{55}}%

\def\eg{\emph{e.g}.} 
\def\ie{\emph{i.e}.}

\def\etal{\emph{et~al}.}

\definecolor{lightgray}{gray}{0.9}

\begin{document}
%
\title{Camouflaged Instance Segmentation In-The-Wild: Dataset, Method, and Benchmark Suite}
%
%
%

\author[1]{Trung-Nghia Le}
\author[2]{Yubo Cao}
\author[4, 3, 8]{Tan-Cong Nguyen}
\author[3, 8]{Minh-Quan Le} 
\author[5, 8]{Khanh-Duy Nguyen}
\author[6]{\\Thanh-Toan Do}
\author[3, 7, 8]{Minh-Triet Tran}
\author[2]{Tam V. Nguyen*\thanks{*Corresponding author. {\it e-mail: tamnguyen@udayton.edu}}\thanks{Manuscript received August 06, 2020; first revised March 21, 2021; second revised August 08, 2021; third revised October 22, 2021; accepted November 17, 2021.}}

\affil[1]{National Institute of Informatics, Japan}
\affil[2]{University of Dayton, United States}
\affil[3]{University of Science, Ho Chi Minh City, Vietnam}
\affil[4]{University of Social Sciences and Humanities, Ho Chi Minh City, Vietnam}
\affil[5]{University of Information Technology, VNU-HCM, Vietnam}
\affil[6]{Monash University, Australia}
\affil[7]{John von Neumann Institute, VNU-HCM, Vietnam} 
\affil[8]{Vietnam National University, Ho Chi Minh City, Vietnam}

\markboth{IEEE TRANSACTIONS ON IMAGE PROCESSING}%
{Camouflaged Instance Segmentation}
%



\setcounter{figure}{-2} 
\makeatletter
\g@addto@macro\@maketitle{
  \begin{figure}[H]
  \setlength{\linewidth}{\textwidth}
  \setlength{\hsize}{\textwidth}
  \centering
    \includegraphics[width=1\textwidth]{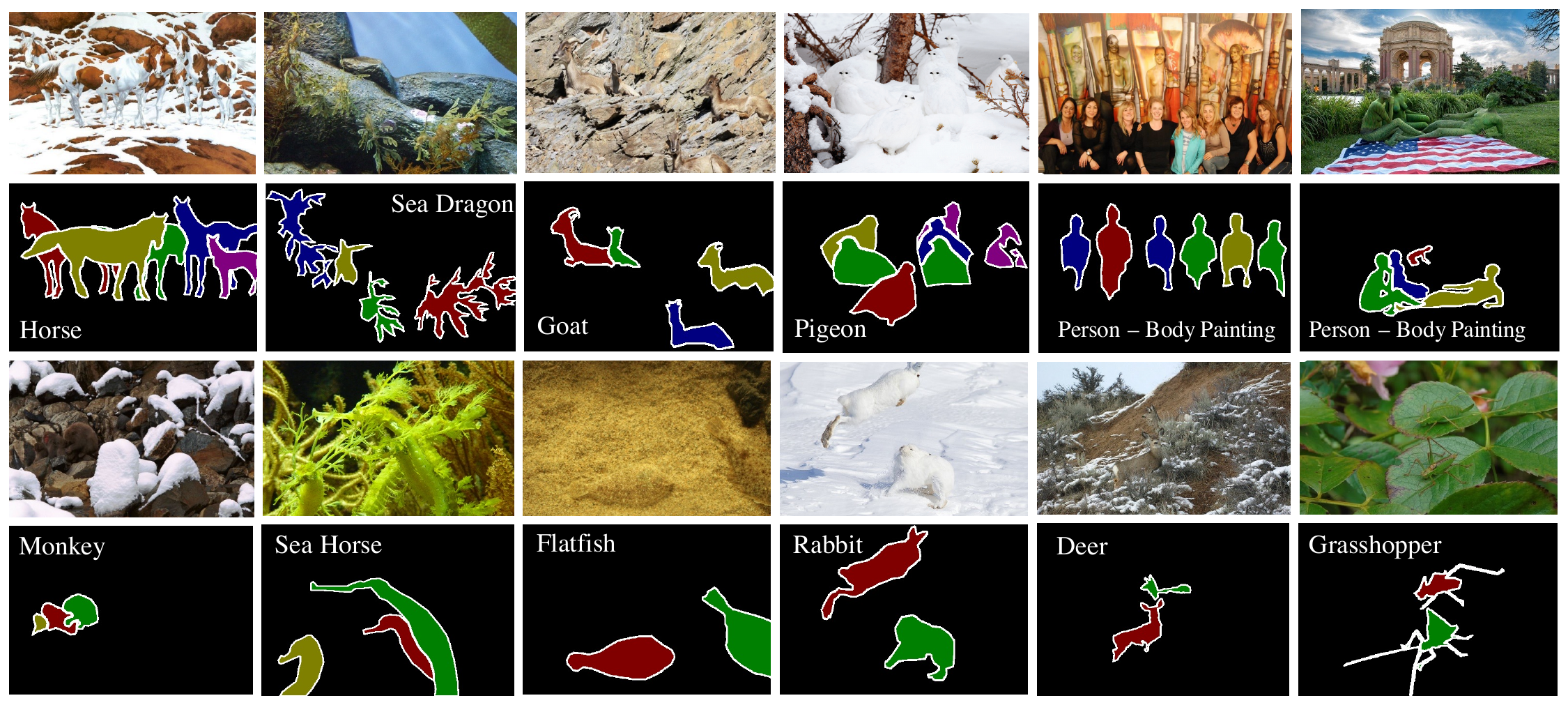}
    \caption{Examples from our \textbf{Cam}ouflaged \textbf{O}bject Plus Plus (CAMO++) dataset with corresponding pixel-level annotations. Plus Plus (++) indicates increments in terms of dataset size and task compared with preliminary CAMO dataset~\cite{ltnghia-CVIU2019}. First and third rows contain original images; second and last rows contain corresponding pixel-wise ground truth images. Each camouflaged instance is represented by distinct color for visualization purposes only. Aim with camouflaged instances is to conceal their texture into the background. Best viewed in color and with zoom.}
    \label{fig:examples}
  \end{figure}
  \vspace{-10mm}
}
\makeatother

\maketitle

\begin{abstract}
This paper pushes the envelope on decomposing camouflaged regions in an image into meaningful components, namely, camouflaged instances. To promote the new task of camouflaged instance segmentation of in-the-wild images, we introduce a dataset, dubbed CAMO++, that extends our preliminary CAMO dataset (camouflaged object segmentation) in terms of quantity and diversity. The new dataset substantially increases the number of images with hierarchical pixel-wise ground truths. We also provide a benchmark suite for the task of camouflaged instance segmentation. In particular, we present an extensive evaluation of state-of-the-art instance segmentation methods on our newly constructed CAMO++ dataset in various scenarios. We also present a camouflage fusion learning (CFL) framework for camouflaged instance segmentation to further improve the performance of state-of-the-art methods. The dataset, model, evaluation suite, and benchmark will be made publicly available on our project page. \footnote{\href{https://sites.google.com/view/ltnghia/research/camo\_plus\_plus}{https://sites.google.com/view/ltnghia/research/camo\_plus\_plus}}
 \end{abstract}
 
\begin{IEEEkeywords}

Camouflaged instance segmentation, in-the-wild image, camouflage dataset, benchmark suite, multimodal learning

\end{IEEEkeywords}

%
\IEEEpeerreviewmaketitle

\section{Introduction} \label{section:introduction}

The term ``camouflage'' was originally used to describe the means that organisms use to disguise their appearance to blend in with their surroundings in order to hunt or avoid being hunted~\cite{Sujit-ICEECS2013}. This natural phenomenon~\cite{ltnghia-CVIU2019} was adopted by humans initially for use on the battlefield. For example, soldiers and military equipment were camouflaged by respectively dressing them and painting it to blend in with the surroundings. This resulted in artificially camouflaged objects~\cite{ltnghia-CVIU2019}. Autonomously identifying camouflaged objects is helpful in various fields of computer vision (search-and-rescue work~\cite{ltnghia-CVIU2019}; wild species discovery and preservation~\cite{ltnghia-CVIU2019}); medical diagnosis (polyp detection and segmentation~\cite{Hemin-ISMICT2019}; COVID-19 infection identification from lung x-rays~\cite{Ferhat-MH2020}); and media forensics (manipulated image/video detection and segmentation~\cite{nhhuy-BTAS2019, ltnghia-ICCV2021}).

Although image segmentation has been worked on for long time, general detectors cannot deal with camouflaged objects~\cite{Kervrann-TIP1995, Boykov-IJCV2006, Li-ICASSP2011, Sulimowicz-ICIP2018}. The detectors initially developed for camouflage detection~\cite{Galun-ICCV2003, Song-ICMT2010, Xue-MTA2016, Pan-MAS2011, Liu-TIP2012, Sengottuvelan-ICETET2008, Yin-PE2011, Gallego-ICIP2014}, which use handcrafted low-level features, are effective only for images with a simple and uniform background. More recently developed deep learning-based detectors~\cite{ltnghia-CVIU2019, Fan-CVPR2020} for camouflaged object segmentation perform only at the region level by mapping each pixel to camouflage/non-camouflage labels, so they do not show the number of camouflaged objects in a scene. None of them targets instance-level segmentation.

In the work reported here, we push the envelope on \textit{decomposing camouflaged regions into meaningful components, namely, camouflaged instances}. A camouflaged object is defined as the set of all camouflaged pixels\footnote{A camouflaged pixel is a pixel belonging to a foreground object that is easily classified as belonging in the background.} in an image without any detailed information such as the number of objects or the semantic meaning~\cite{ltnghia-CVIU2019}. In contrast, a \textit{camouflaged instance consists of only meaningful pixels that cover an object instance}. 

Camouflaged instance segmentation is more challenging than conventional camouflaged object segmentation in the sense that it \textit{not only maps each pixel to a label but also assigns an instance identity to each pixel}. To the best of our knowledge, this is the first work to address camouflaged instance segmentation. Existing camouflaged object segmentation methods were developed under the assumption that camouflaged objects are always present in an image~\cite{Fan-CVPR2020, Lamdouar-ACCV2020, Jinchao-AAAI2021}. However, this assumption is not always satisfied in practice. In contrast, \textit{our work focuses on segmentation of in-the-wild camouflage images without any assumption}. To simulate the real-world, we aim to \textit{segment camouflaged instances in unrestricted images, meaning that camouflaged instances are not always present}.

To this end, we introduce a dataset designed explicitly for the task of camouflaged instance segmentation. The dataset contains 5,500 images of people and more than 90 animal species with hierarchically pixel-wise annotation for all images. There are both camouflage and non-camouflage images at a ratio of approximately 50:50. It can thus serve as a benchmark for not only the camouflaged instance segmentation task but also the conventional camouflaged object segmentation task. We also provide a benchmark suite to facilitate the evaluation and advance the task of camouflaged instance segmentation. Particularly, we evaluate and analyze state-of-the-art instance segmentation methods in various scenarios. Note that previous work~\cite{Fan-CVPR2020, Lamdouar-ACCV2020, Jinchao-AAAI2021, ltnghia-AAAI2021} used only camouflage images to train and evaluate their methods under the assumption that camouflaged objects are always present an image. In addition, existing camouflage datasets~\cite{ltnghia-CVIU2019, Fan-CVPR2020} do not have the ground truth of non-camouflage images. This does not correspond to the real world since an animal or person is considered to be camouflaged depending on the surrounding context. Both camouflage and non-camouflage image in our dataset are annotated, enabling both image types to be used for simulating the real world. In addition, we present a fusion method that further improves the performance of the state-of-the-art methods.

\begin{figure*}[t]
    \centering
    \centering
    \begin{tabularx}{\linewidth}{*{3}{X}}
        \hfill\includegraphics[width=1\linewidth]{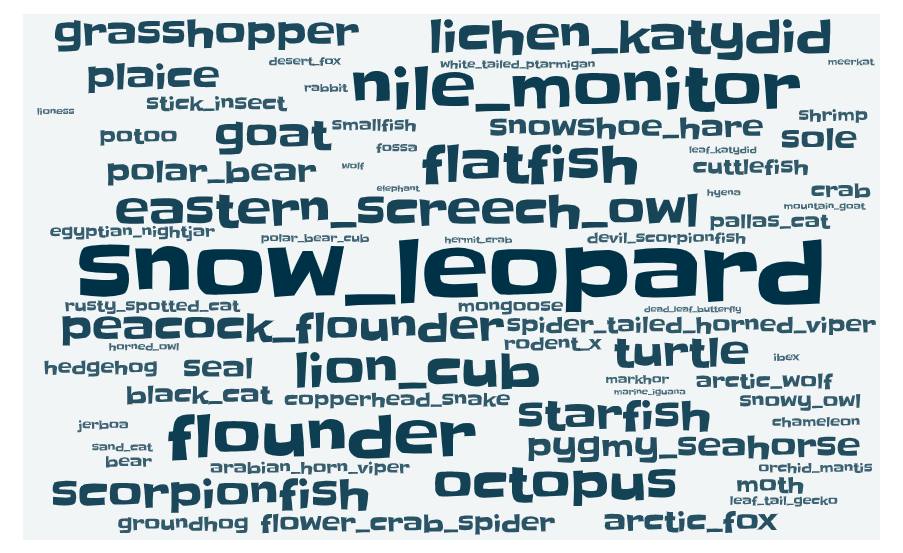}\hspace*{\fill} & \hfill\includegraphics[width=1\linewidth]{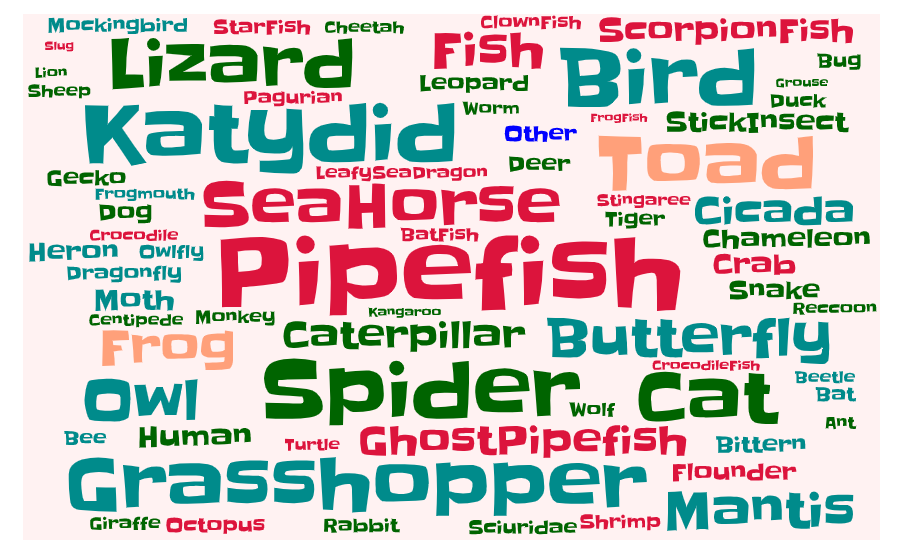}\hspace*{\fill} & \hfill\includegraphics[width=1\linewidth]{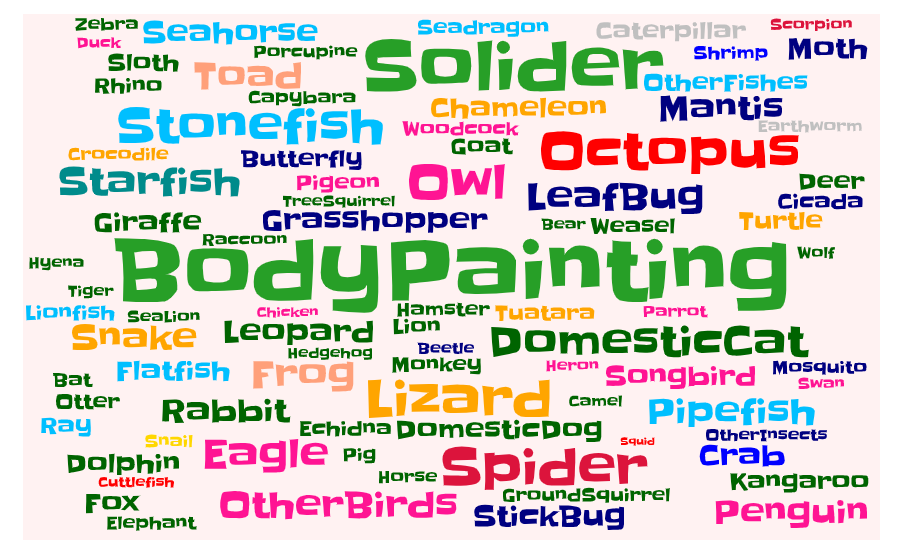}\hspace*{\fill} \\
        \centering \footnotesize{a) MoCA~\cite{Lamdouar-ACCV2020}} &
        \centering \footnotesize{b) COD~\cite{Fan-CVPR2020}} & 
        \centering \footnotesize{c) CAMO++}
    \end{tabularx}
    \caption{Word cloud of category distribution of camouflaged instances. Colors represent different meta-categories. Our CAMO++ dataset has 12 meta-categories while COD dataset has only 5 meta-categories.}
    \label{fig:word_cloud}
\end{figure*}

We summarize the contributions of this paper as follows: 
\begin{itemize}

\item We address a new task of \textbf{camouflaged instance segmentation in-the-wild} and analyze in depth the challenges of performing this task. To the best of our knowledge, this is the first work defining and exploring camouflaged instance segmentation. Finding camouflaged instances in a scene is a useful task, and it should be an interesting problem for the computer vision community. A few methods have been reported that can perform camouflaged object segmentation, but none can perform instance-level segmentation. In addition, they are based on the assumption that camouflaged objects are always present in an image; in contrast, our proposed method performs segmentation on unrestricted images without any assumption.

\item We present a new image dataset to promote the task of camouflaged instance segmentation. Our newly constructed \textbf{Cam}ouflaged \textbf{O}bject Plus Plus (\textbf{CAMO++}) dataset, extended from our preliminary CAMO dataset~\cite{ltnghia-CVIU2019}, consists of $5,500$ images of people and more than 90 animal species, with 2700 camouflage images and 2800 non-camouflage images. All images are hierarchically annotated with meta-category labels, fine-category labels, bounding boxes, and instance-level masks. Pixel-wise ground truths were manually annotated for all instances in each image. 

\item We provide a benchmark suite for the camouflaged instance segmentation task. In particular, we present the results of an extensive evaluation of state-of-the-art instance segmentation methods in various scenarios. We further provide an in-depth analysis of the experiments. The CAMO++ dataset, evaluation suite, and benchmark will be made available on our website along with paper publication\footnote{\href{https://sites.google.com/view/ltnghia/research/camo\_plus\_plus}{https://sites.google.com/view/ltnghia/research/camo\_plus\_plus}}.

\item We present a camouflage fusion learning (CFL) framework for camouflaged instance segmentation that leverages the advantages of state-of-the-art methods in camouflaged instance segmentation.

\end{itemize}

\begin{table*}[t]
\caption{Statistics of camouflage datasets.}
\resizebox{1\textwidth}{!}{
\begin{tabular}{l|c|c|c|c|c|c|c|c|c|c}
\toprule
\textbf{Dataset} & \textbf{Year} & \textbf{Publication} & \textbf{Type} & \textbf{\#Img.} & \textbf{\#Ann. Img.} & \textbf{\#Obj. Cat.} & \textbf{\textcolor{red}{\#Ins.} or \#Obj.} & \textbf{Bbox. GT} & \textbf{Obj. Mask GT} & \textbf{Ins. Mask GT} \\
\midrule
CamouflagedAnimals~\cite{Bideau-ECCV2016} & 2016 & ECCV & Video & 839 & 181 & 6 & \textcolor{red}{224} & \xmark & \cmark & \cmark \\
CHAMELEON~\cite{Skurowski-2018} & 2018 & - & Image & 76 & 76 & - & 76 & \xmark & \cmark & \xmark \\
CAMO~\cite{ltnghia-CVIU2019} & 2019 & CVIU & Image & 2,500 & 1,250 & 8 & 1,250 & \xmark & \cmark & \xmark \\
COD~\cite{Fan-CVPR2020} & 2020 & CVPR & Image & 10,000 & 5,066 & 69 & \textcolor{red}{5,930} & \cmark & \cmark & \cmark \\
MoCA~\cite{Lamdouar-ACCV2020} & 2020 & ACCV & Video & 37,250 & 7,617 & 67 & 7,617 & \cmark & \xmark & \xmark \\
\midrule
\rowcolor{lightgray} CAMO++ & 2020 & - & Image & 5,500 & 5,500 & 93 & \textcolor{red}{32,756} & \cmark & \cmark & \cmark \\
\bottomrule
\end{tabular}
}
\label{table:camouflage_datasets}
\end{table*}

The remainder of this paper is organized as follows. Section~\ref{section:related_work} summarizes related work. Next, Section~\ref{section:dataset} introduces the newly constructed CAMO++ dataset. Section~\ref{section:method} presents our proposed CFL framework for camouflaged instance segmentation. Section~\ref{section:benchmark} presents the benchmark suite and the results of our evaluation of baselines on the newly constructed dataset. Finally, Section~\ref{section:conclusion} summarizes the key points and mentions future work.

\section{Related Work} \label{section:related_work}

\subsection{Camouflaged Object Segmentation}

When a large-enough area in a foreground object can be easily classified as background, the pixels in that area can be considered to be camouflaged. As mentioned above, a camouflaged object is defined as the set of all camouflaged pixels in an image without any further detailed information such as the number of objects or the semantic meaning~\cite{ltnghia-CVIU2019}. Although camouflaged object recognition has a wide range of applications, this research field has not been well explored in the literature. Early work related to camouflage detection focused on the foreground region even when some of its texture was similar to the background~\cite{Galun-ICCV2003, Song-ICMT2010, Xue-MTA2016}. The foreground was distinguished from the background on the basis of simple features, such as color, intensity, shape, orientation, and edge. A few methods based on handcrafted low-level features have been presented for tackling the problem of camouflage detection~\cite{Pan-MAS2011, Liu-TIP2012, Sengottuvelan-ICETET2008, Yin-PE2011, Gallego-ICIP2014}. However, they are effective only for images with a simple and uniform background. Thus, their performances are unsatisfactory in camouflaged object segmentation due to the substantial similarity between the foreground and the background. 

Recently, Le \etal~\cite{ltnghia-CVIU2019} proposed an end-to-end network, dubbed ANet, for camouflaged object segmentation through integrating classification information into segmentation. The idea of utilizing classification for segmentation can be helpful when appropriately applied to multiple-region segmentation. Following the same direction, Fan \etal~\cite{Fan-CVPR2020} subsequently developed SINet, which includes two main modules, namely a search module and an identification module. This network is based on simulated hunting, in which a predator first judges whether a potential prey exists; \ie, it searches for prey. Once a target animal is identified, it can be caught. Yan \etal~\cite{Jinnan-IEEEAccess2021} recently introduced MirrorNet, a dual-stream network comprising a main stream and a mirror stream. This bio-inspired network effectively captures different layouts of the scene and thereby boosts segmentation accuracy. Jinchao \etal~\cite{Jinchao-AAAI2021} presented the TINet, which interactively refines multi-level texture and segmentation features and thereby gradually enhances the segmentation of camouflaged objects. To the best of our knowledge, there has been no previous work on camouflaged instance segmentation.

\subsection{Camouflage Datasets}

Table~\ref{table:camouflage_datasets} summarizes the main characteristics of different datasets in camouflage research. CamouflagedAnimals~\cite{Bideau-ECCV2016} and CHAMELEON~\cite{Skurowski-2018} were the first two camouflage datasets with mask annotations. However, they do not contain enough images to train deep learning methods. Le \etal~\cite{ltnghia-CVIU2019} created the CAMO dataset, the first camouflage dataset with more than 1000 annotated images. It contains 1250 annotated images, which is a limited number of samples to train and evaluate deep learning methods. Fan \etal~\cite{Fan-CVPR2020} subsequently created the COD dataset, which comprises 10,000 images (both camouflage and non-camouflage). However, they annotated only 5,000 camouflage images. Lamdouar \etal~\cite{Lamdouar-ACCV2020} recently developed the MoCA dataset for the camouflage object detection task; it contains only bounding box ground truths. Although CAMO~\cite{ltnghia-CVIU2019} and COD~\cite{Fan-CVPR2020} were constructed with multiple levels of ground truth, they have been used only for the camouflaged object-level segmentation problem. To the best of our knowledge, our newly constructed CAMO++ dataset is the first dataset fully supporting camouflaged instance-level segmentation.

\subsection{Instance Segmentation}

Instance segmentation is the task of unifying object detection and semantic segmentation. It has been intensively studied in recent years using either the segmentation-based approach or the proposal-based approach. With the segmentation-based approach~\cite{Kirillov-CVPR2017, Levinkov-CVPR2017, Liang-TPAMI2017, Zhang-CVPR2016, Konstantin-ICCV2019}, two-stage processing is generally used: segmentation first and then instance clustering. With the proposal-based approach~\cite{Dai-CVPR2016, Kaiming-ICCV2017, Yi-CVPR2017}, on the other hand, bounding boxes are first predicted and then parsed to obtain mask regions~\cite{Dai-CVPR2016}, or an object detection model is used (\eg, Faster RCNN~\cite{Ren-NeurIPS2015} or R-FCN~\cite{Dai-NeurIPS2016}) to classify mask regions~\cite{Kaiming-ICCV2017, Yi-CVPR2017}. Proposal-based methods, which achieve state-of-the-art performance, have gained popularity due to their superiority over segmentation-based methods. Hence, this paper solely focuses on proposal-based methods.

Proposal-based methods can be categorized into single-stage and two-stage processes. The two-stage methods detect and then segment: they first perform object detection to extract a bounding box around each instance object and then perform binary segmentation inside each bounding box to separate the foreground (object) and the background. Two-stage methods (\eg, Mask RCNN~\cite{Kaiming-ICCV2017} and its variants) are quite slow and thus are not practical for many real-time applications. Mask RCNN~\cite{Kaiming-ICCV2017}, the first end-to-end model for instance segmentation, is an extension of Faster RCNN~\cite{Ren-NeurIPS2015}:  a branch was added for predicting an object mask in parallel with the existing branch for bounding box detection. Mask Scoring RCNN (MS RCNN)~\cite{Huang-CVPR2019}, Cascade Mask RCNN~\cite{Zhaowei-CVPR2018}, and PANet~\cite{Liu-CVPR2018} are extensions of Mask RCNN that improve the quality of segmented instances. MS RCNN~\cite{Huang-CVPR2019} contains a network block on top of Mask RCNN that learns the quality of the predicted instance masks. Cascade Mask RCNN~\cite{Zhaowei-CVPR2018}, a multi-stage architecture, consists of a sequence of detectors trained with increasing intersection over union (IoU) thresholds that are sequentially more selective against close false positives. PANet~\cite{Liu-CVPR2018} is aimed at boosting information flow in the feature extractor through bottom-up path augmentation.

The single-stage methods were inspired by anchor-free object detection methods (such as CenterNet~\cite{Zhou-2019} and FCOS~\cite{Tian-ICCV2019}). Generally, these methods are faster than two-stage methods. Some can even run in real time. YOLACT~\cite{Bolya-ICCV2019} is one of the first such methods attempting real-time instance segmentation. YOLACT breaks instance segmentation into two parallel subtasks (\ie, generating a set of prototype masks and predicting per-instance mask coefficients) and then linearly combines the prototypes with the mask coefficients. BlendMask~\cite{Chen-CVPR2020} and CenterMask~\cite{Lee-CVPR2020}, which were extended from YOLACT, are aimed at blending cropped prototype masks with a finer-grained mask within each bounding box. CenterMask~\cite{Lee-CVPR2020} adds a spatial attention-guided mask branch to an anchor-free single-stage object detector (FCOS~\cite{Tian-ICCV2019}). BlendMask~\cite{Chen-CVPR2020} first predicts dense per-pixel position-sensitive instance features with very few channels and then merges the attention map for each instance through a blender module. PolarMask~\cite{Xie-CVPR2020} performs instance segmentation by predicting the contour of each instance by instance center classification and dense distance regression in a polar coordinate. EmbedMask~\cite{Ying-CVPR2020} generates embeddings for pixels and proposals to assign pixels to the mask of the proposal if their embeddings are similar. TensorMask~\cite{Chen-ICCV2019} performs dense sliding window instance segmentation using structured 4D tensors to represent masks over a spatial domain. RetinaMask~\cite{Fu-2019} adds a novel instance mask prediction head to the single-shot RetinaNet~\cite{Lin2-CVPR2017} detector. FCIS~\cite{Yi-CVPR2017} and CondInst~\cite{Tian-ECCV2020} use fully convolutional networks to produce masks. SOLO~\cite{Wang-ECCV2020} reformulates instance segmentation as category prediction and mask generation and directly outputs masks without computing bounding boxes.  

To the best of our knowledge, this is the first work to address camouflaged instance segmentation. Given the lack of a large-scale dataset for training and testing purposes, we created a benchmark for the task of camouflaged instance segmentation by training instance segmentation methods on our newly constructed CAMO++ dataset. We further conducted an extensive evaluation of state-of-the-art instance segmentation methods in various scenarios. 


\begin{figure}[t!]
    \centering
         \includegraphics[width=1\linewidth]{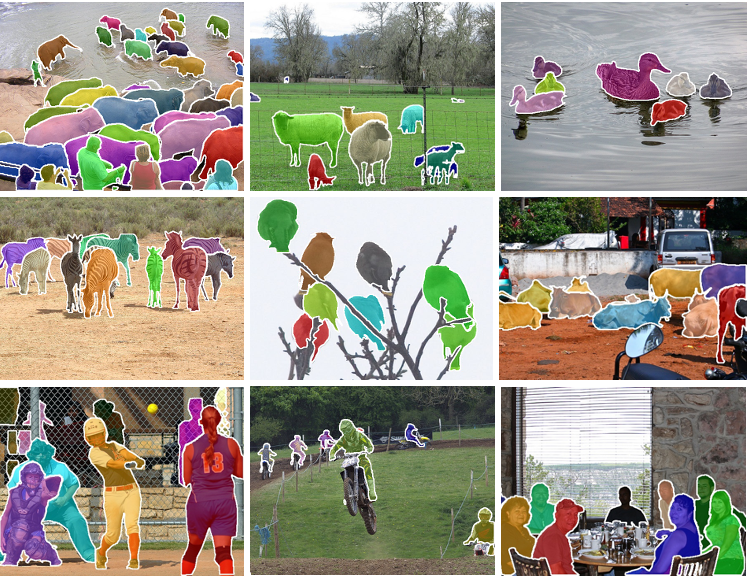}
    \caption{Examples of non-camouflage images. Each instance is overlaid with a distinctive color for visualization purposes only. Best viewed in color with zoom.}
    \label{fig:non_cam_examples}
\end{figure}

\begin{table}[t!]
\caption{Distribution of data in CAMO++ dataset.}
\centering
\begin{tabular}{l|c|c|c}
\toprule
 & Training & Testing & Total \\
 \midrule
 Camouflage Images & 1,700 & 1,000 & 2,700 \\
 Non-camouflage Images & 1,800 & 1,000 & 2,800 \\
 Instances & 17,735 & 15,021 & 32,756 \\
\bottomrule
\end{tabular}
\label{table:data}
\end{table}

\begin{figure*}[!t]
    \centering
    \begin{tabularx}{\linewidth}{*{2}{X}}
        \hfill\includegraphics[width=1\linewidth]{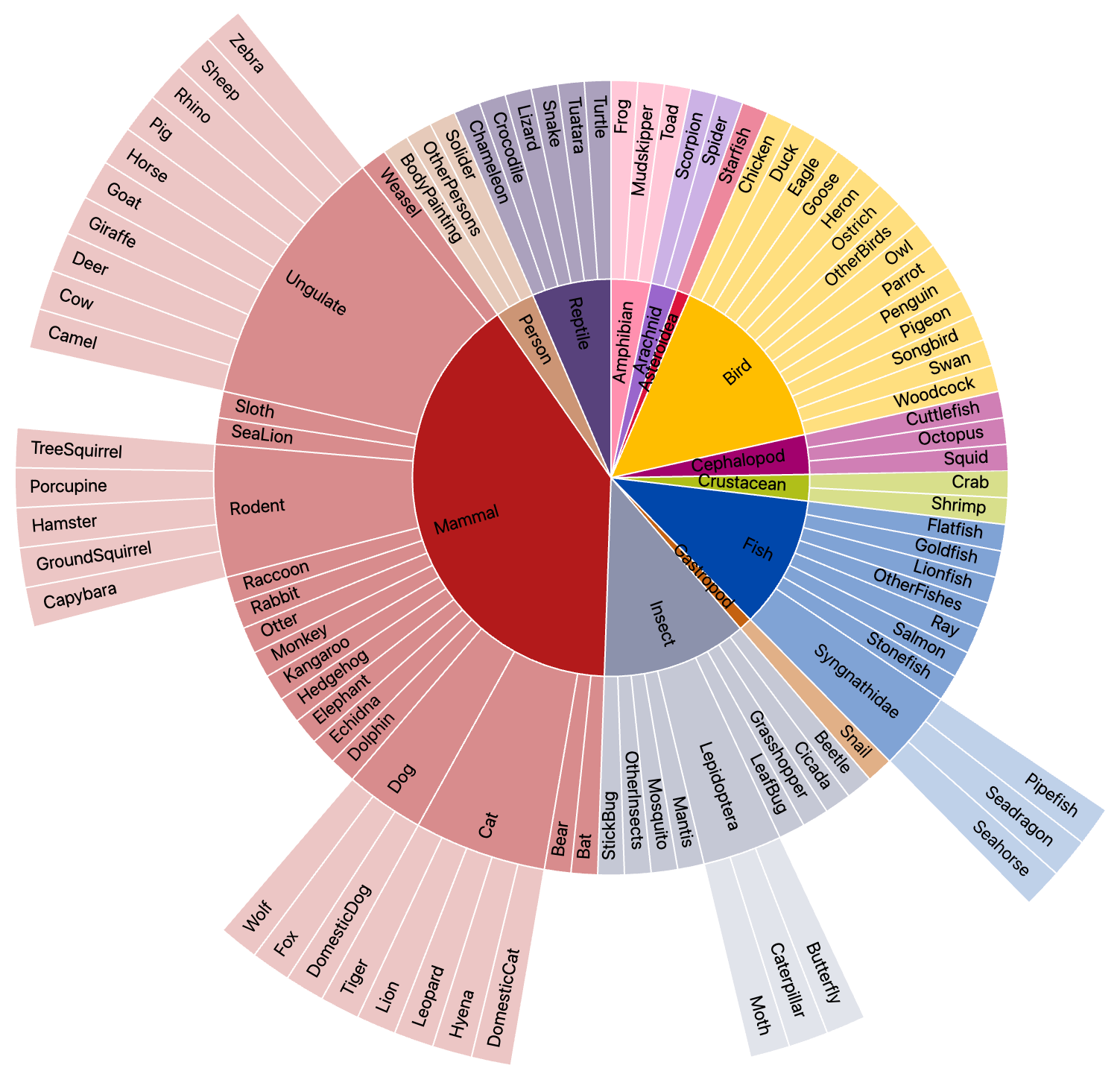}\hspace*{\fill} & 
        \hfill\includegraphics[width=1\linewidth]{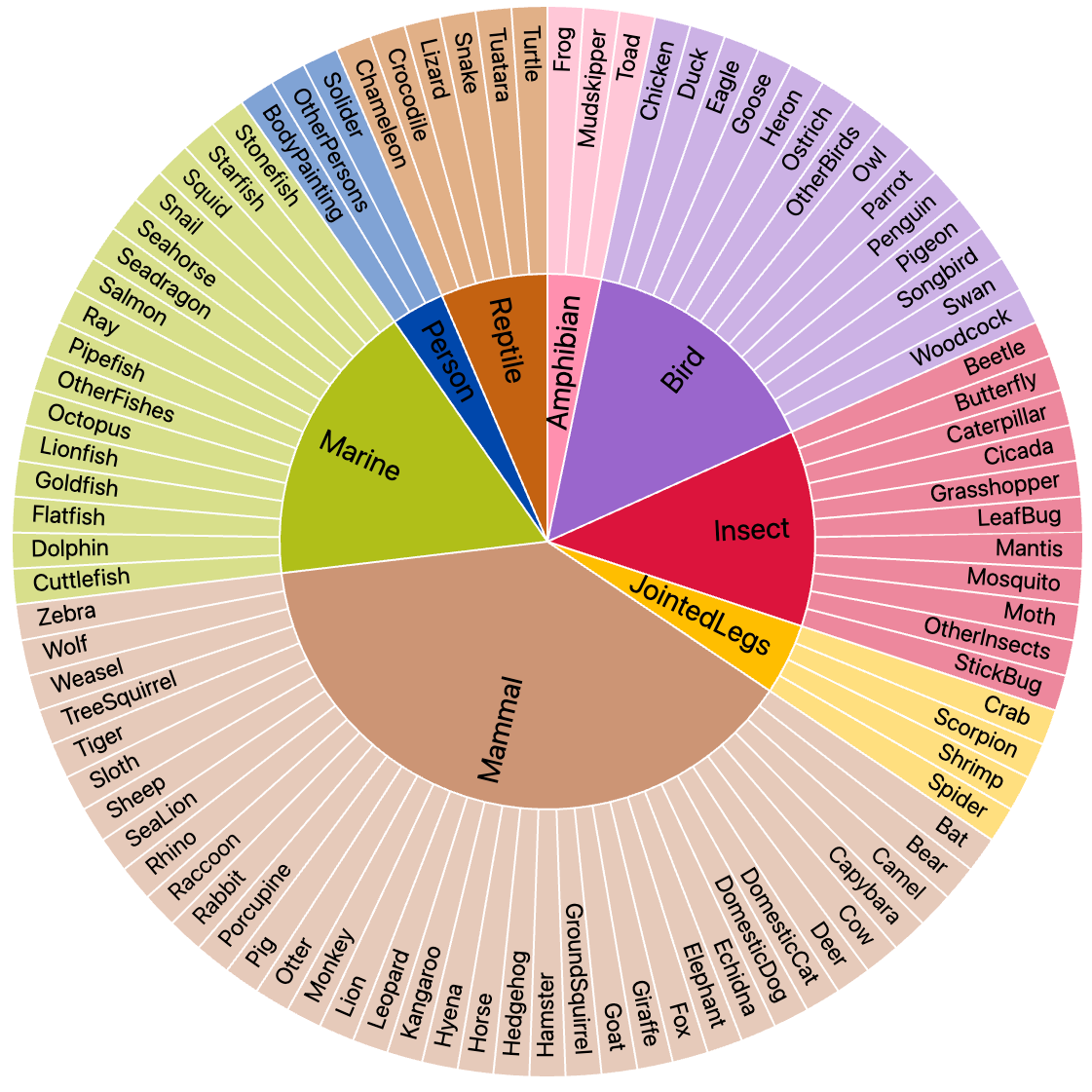}\hspace*{\fill} \\
        \centering \footnotesize{a) Biology taxonomic structure.} &
        \centering \footnotesize{b) Vision taxonomic structure.} 
    \end{tabularx}
    \caption{Hierarchical taxonomic structure of our CAMO++ dataset.}
    \label{fig:category}
\end{figure*}

\begin{figure}[!t]
    \centering
    \begin{tabularx}{\linewidth}{*{2}{X}}
        \hfill\includegraphics[width=1\linewidth]{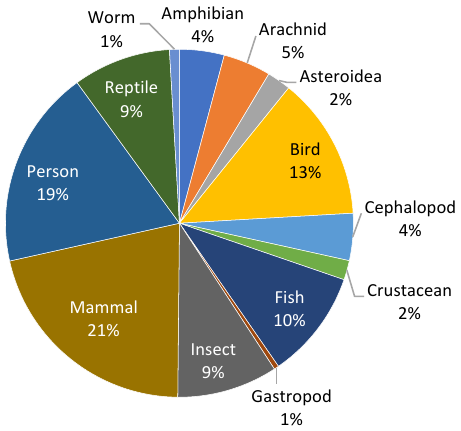}\hspace*{\fill} & 
        \hfill\includegraphics[width=.95\linewidth]{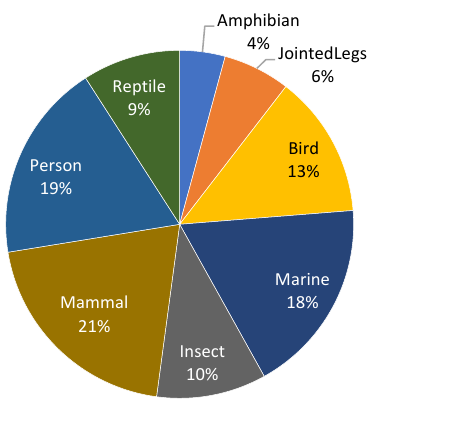}\hspace*{\fill} \\
        \centering \footnotesize{a) Biology meta-categories 
        } &
        \centering \footnotesize{b) Vision meta-categories
        } 
    \end{tabularx}
    \caption{Distribution of camouflage meta-categories in CAMO++ dataset.}
    \label{fig:category_img}
\end{figure}

\begin{figure}[!t]
    \centering
    \includegraphics[width=\linewidth]{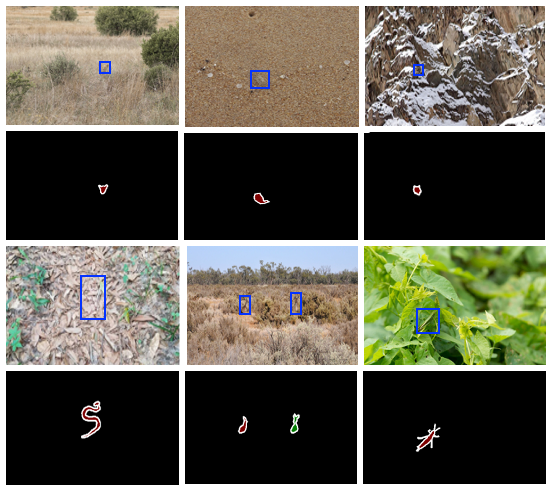}
    \caption{Examples of tiny camouflaged instances (best viewed online in color with zoom). Each instance is overlaid with a distinctive color for visualization purposes only.}
    \label{fig:tiny_ins}
\end{figure}

\section{Camouflaged Instance Dataset}
\label{section:dataset}

The core contribution of this paper is our CAMO++ dataset, which extends our preliminary CAMO dataset~\cite{ltnghia-CVIU2019}. This new large-scale dataset enables us to train and evaluate state-of-the-art methods for the task of camouflaged instance segmentation.

\subsection{Dataset Construction}

\subsubsection{Camouflage Image Collection}

We initially collected 4000 images containing at least one camouflaged object. We collected them from the Internet using various search terms combining an adjective (``camouflaged,'' ``concealed,'' ``hidden'') with an animal name (\eg, cat, dog, seahorse), a person-related term (\eg, soldier, body-painting), and/or an environment (\eg, marine, underwater, mountain, desert, forest).

We manually discarded images with low resolution. We mixed the remaining images with the 1250 camouflage images in our preliminary CAMO dataset~\cite{ltnghia-CVIU2019} and manually discarded the duplicates. We ended up with 2700 camouflage images.

We asked ten annotators to identify the camouflaged instances in each image and annotate them using a custom-designed interactive segmentation tool. It took each annotator 5–20 minutes to annotate an image depending on its complexity. The annotation stage thus spanned a few months. The outcome of this process was a binary mask for each image and the hierarchical category for each instance (see Figure~\ref{fig:examples}).

\begin{figure*}[!t]
    \centering
    \begin{tabularx}{\linewidth}{*{3}{X}}
        \hfill\includegraphics[width=1\linewidth]{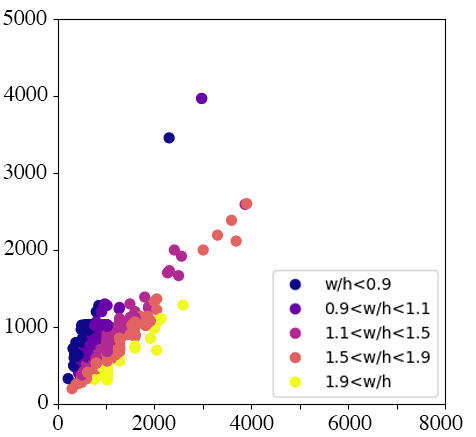}\hspace*{\fill} & 
        \hfill\includegraphics[width=1\linewidth]{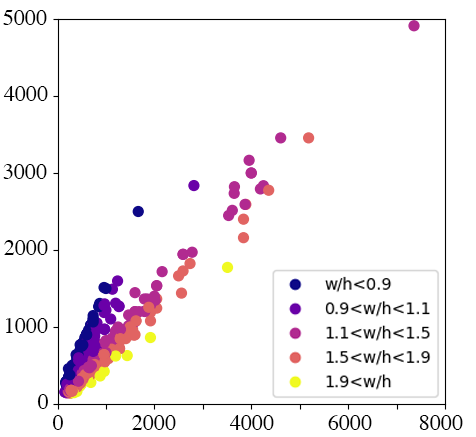}\hspace*{\fill} & 
        \hfill\includegraphics[width=1\linewidth]{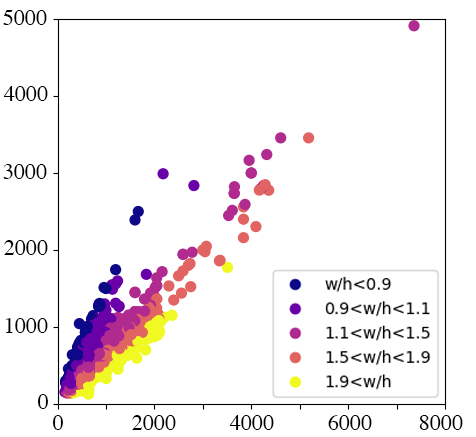}\hspace*{\fill} \\
        \centering \footnotesize{a) COD~\cite{Fan-CVPR2020}} & \centering  \footnotesize{b) CAMO~\cite{ltnghia-CVIU2019}} &
        \centering \footnotesize{c) CAMO++} 
    \end{tabularx}
    \caption{Image dimension distribution (w=width; h=height). Please zoom in for viewing ease.}
    \label{fig:image_resolution}
\end{figure*}

\begin{figure*}[!t]
    \centering
    \begin{tabularx}{0.9\linewidth}{*{2}{X}}
        \hfill\includegraphics[width=1\linewidth]{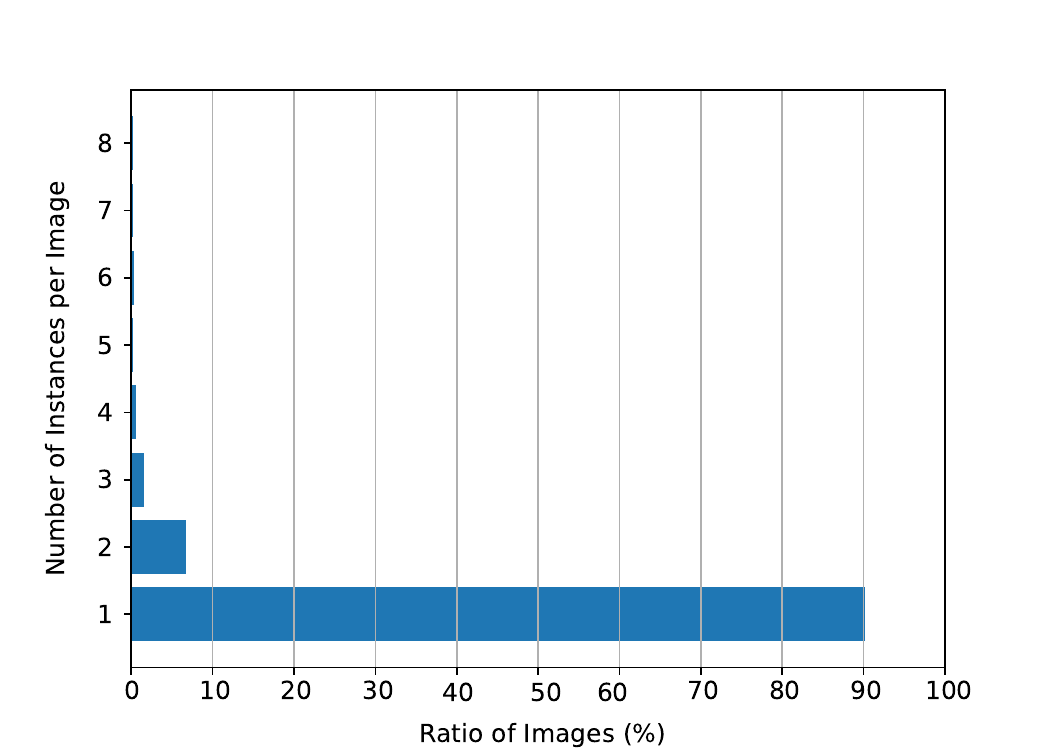}\hspace*{\fill} &
        \hfill\includegraphics[width=1\linewidth]{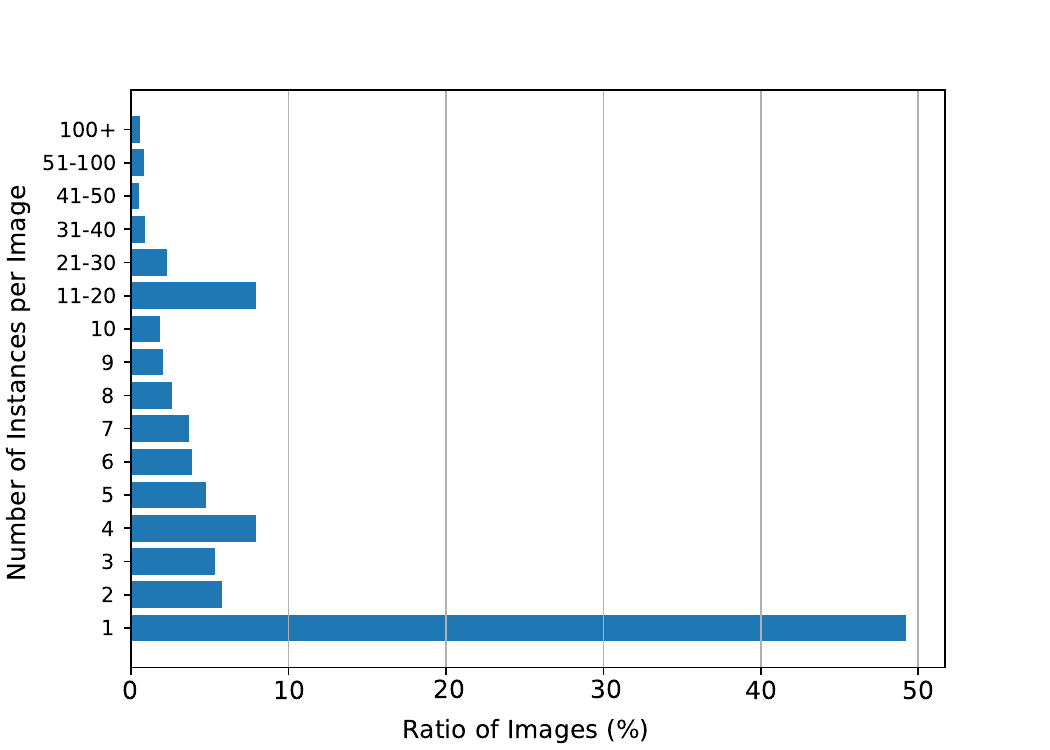}\hspace*{\fill} \\
        \centering \footnotesize{a) COD~\cite{Fan-CVPR2020}} &
        \centering \footnotesize{b) CAMO++} 
    \end{tabularx}
    \caption{Instances per image distribution.}
    \label{fig:ins-distribution}
\end{figure*}

\subsubsection{Non-Camouflage Image Collection}

We manually selected 2800 images from the large vocabulary instance segmentation (LVIS) dataset~\cite{Gupta-CVPR2019} that contained at least one human or animal instance. We manually selected the images to ensure that they did not contain any camouflaged instances, which could result in false-positive segmentation. Figure \ref{fig:non_cam_examples} shows examples of the non-camouflage images. 

\subsubsection{Dataset Splits}
\label{section:dataset_splits}

We randomly split the 2700 camouflage images into separate training and testing sets: the training set consisted of 1700 images, and the testing set consisted of 1000 images. We also randomly split the 2800 non-camouflage images into training and testing sets: 1800 images and 1000 images, respectively.

\begin{figure*}[t]
    \centering
    \centering
    \begin{tabularx}{\linewidth}{*{4}{X}}
        \hfill\includegraphics[width=1\linewidth]{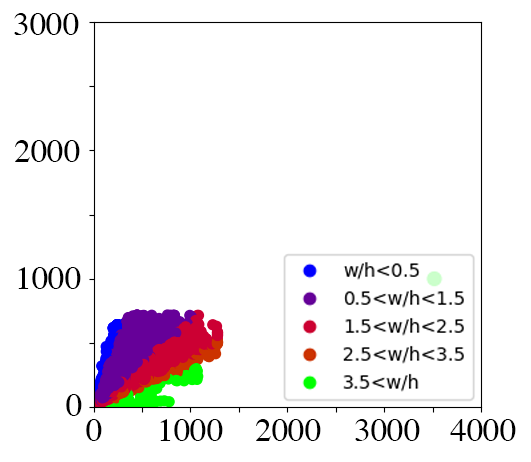}\hspace*{\fill} & \hfill\includegraphics[width=1\linewidth]{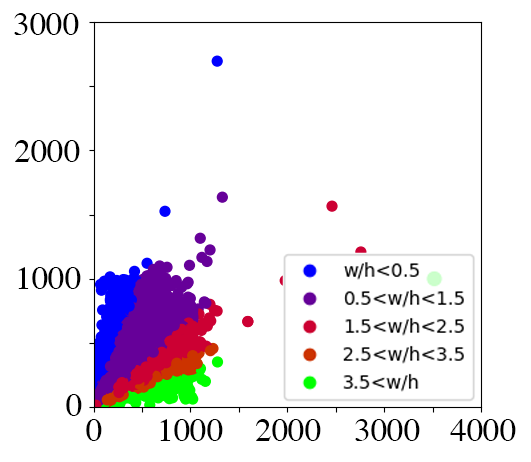}\hspace*{\fill} &
         \hfill\includegraphics[width=1\linewidth]{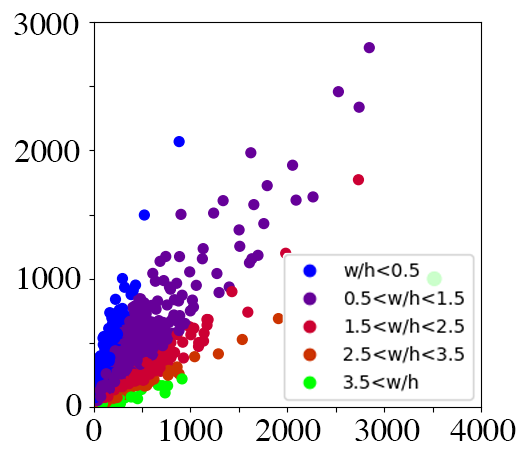}\hspace*{\fill} &
        \hfill\includegraphics[width=1\linewidth]{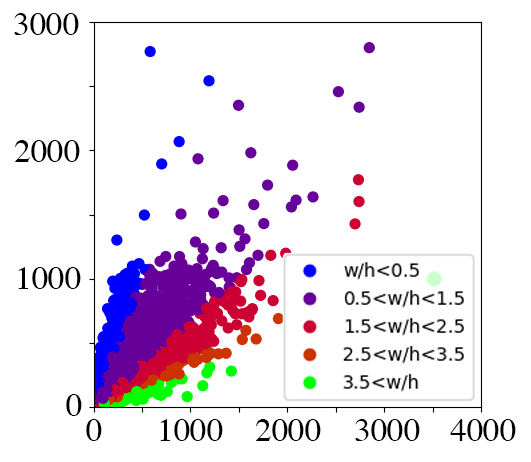}\hspace*{\fill} \\
        \centering \footnotesize{a) MoCA~\cite{lamdouar2020betrayed}} &
        \centering \footnotesize{b) COD~\cite{Fan-CVPR2020}} &
        \centering \footnotesize{c) CAMO~\cite{ltnghia-CVIU2019}} &
        \centering \footnotesize{d) CAMO++} 
    \end{tabularx}
    \caption{Instance bounding box distribution of four camouflage datasets (w=width; h=height) (best viewed online in color with zoom).}
    \label{fig:bbox-distribution}
\end{figure*}

\begin{figure*}[t]
    \centering
    \begin{tabularx}{\linewidth}{*{4}{X}}
        \hfill\includegraphics[width=1\linewidth]{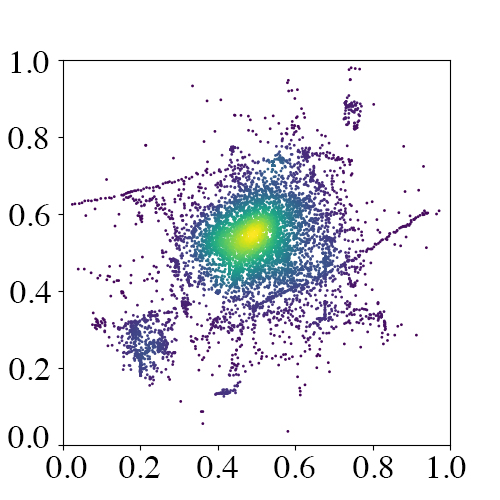}\hspace*{\fill} & \hfill\includegraphics[width=1\linewidth]{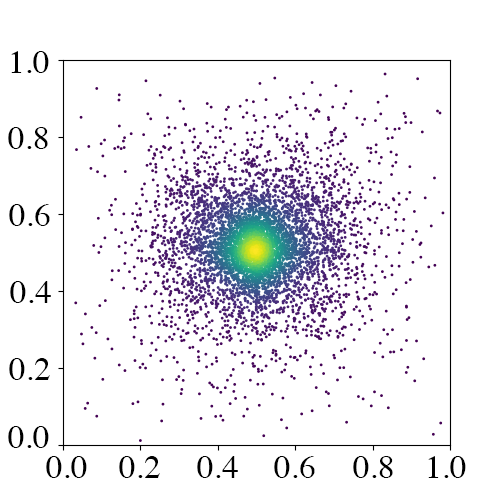}\hspace*{\fill} & 
        \hfill\includegraphics[width=1\linewidth]{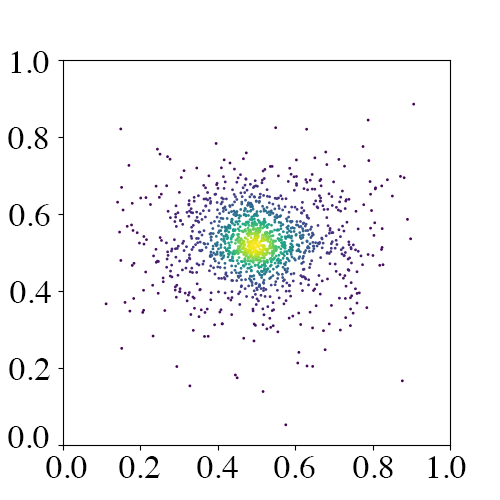}\hspace*{\fill} & 
        \hfill\includegraphics[width=1\linewidth]{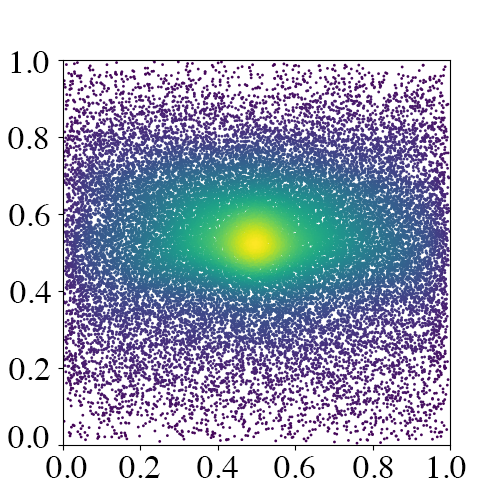}\hspace*{\fill} \\
        \centering \footnotesize{a) MoCA~\cite{lamdouar2020betrayed}} & \centering \footnotesize{b) COD~\cite{Fan-CVPR2020}} &
        \centering  \footnotesize{c) CAMO~\cite{ltnghia-CVIU2019}} &
        \centering \footnotesize{d) CAMO++} 
    \end{tabularx}
    \caption{Instance center bias of four datasets (best viewed online in color with zoom-in).}
    \label{fig:center_bias}
\end{figure*}

\begin{figure}[!t]
    \centering
    \includegraphics[width=\linewidth]{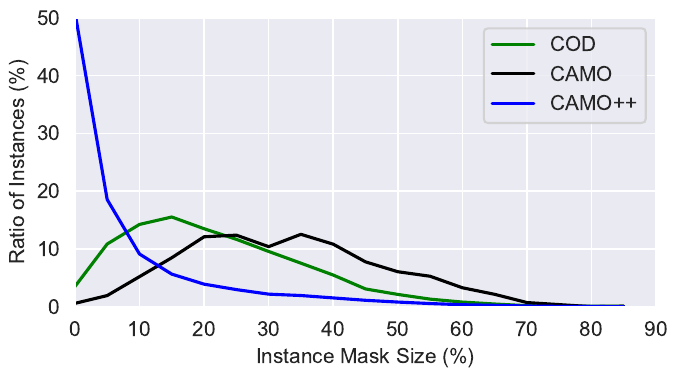}
    \caption{Instance mask size distribution. CAMO++ dataset contains the most small objects. Mask size of camouflaged instances in CAMO++ dataset is the most diverse (\ie, tiny, small, medium, large).}
    \label{fig:ins-size}
\end{figure}

\subsection{Dataset Specifications and Statistics}

We describe in this section the \textbf{Camouflaged Object Plus Plus (CAMO++)} dataset designed explicitly for the task of camouflaged instance segmentation. The \textbf{++} indicates the increments in terms of dataset size and task compared with our preliminary CAMO dataset~\cite{ltnghia-CVIU2019}. Example images are shown in Figure~\ref{fig:examples} along with the corresponding ground-truth label annotations. As shown in Table~\ref{table:camouflage_datasets}, the CAMO++ dataset has the most object categories and object instances. Note that the COD dataset~\cite{Fan-CVPR2020} provides the ground truth only for camouflaged instances that do not satisfy the in-the-wild setting in nature.

\textbf{Category Diversity.} 
The CAMO++ dataset is focused on various kinds of camouflaged pieces as shown by the biology-inspired hierarchical categorization illustrated in Figure \ref{fig:category}a. The CAMO++ dataset consists of 13 biological meta-categories (\ie, amphibian, arachnid, asteroidea, bird, cephalopod, crustacean, fish, gastropod, insect, mammal, person, reptile, and worm) and 93 categories. Each category contains 352 instances on average. Figure \ref{fig:category_img}a shows the ratios of the 13 biological meta-categories in the CAMO++ dataset: each biology meta-category contains 2520 instances on average. The distributions of camouflaged instances by category in three datasets are represented by the word clouds in Figure~\ref{fig:word_cloud}.

Since biological categorization can be difficult for machines and computer vision experts to understand, we re-organized the categories on the basis of vision features (\ie, appearance) combined with biological features (\ie, behavior, living environment). We created a new set of eight meta-categories (\ie, amphibian, bird, insect, jointed legs, mammal, marine, person, reptile). Figure~\ref{fig:category_img}b shows the ratios of the eight vision meta-categories in the CAMO++ dataset: each vision meta-category contains 4095 instances on average.

Our CAMO++ dataset contains more object categories and meta-categories than previous datasets. It contains 93 categories and 13 meta-categories in comparison with the 69 categories and 5 meta-categories of the COD dataset~\cite{Fan-CVPR2020} and the 67 categories of the MoCA dataset~\cite{Lamdouar-ACCV2020} (see Table~\ref{table:camouflage_datasets}). The diversity of the CAMO++ dataset should enable a comprehensive understanding of camouflage instances in both the biology and computer vision communities.

\textbf{Image Dimension.}
As shown in Figure~\ref{fig:image_resolution}, the CAMO++ dataset has the greatest diversity in image dimensions. Moreover, it has more high-resolution images than the COD dataset.

\textbf{Instance Density.} 
In the CAMO++ dataset, each image has from 1 to more than 100 instances. It has 6.0 instances per image on average, whereas the COD dataset has only 1.2 instances per image on average and a maximum of 8. As illustrated in Figure \ref{fig:ins-distribution}, the CAMO++ dataset has a large number of images with multiple instances, including separate single instances and spatially connected/overlapping instances. They account for $51\%$ of the images: $38\%$ have from 2 to 10 instances, $10\%$ have from 11 to 30 instances, and the remaining $3\%$ have more than 30 instances. In contrast, only $10\%$ images in the COD dataset have multiple singe instances. This makes our CAMO++ dataset more challenging for the camouflaged instance segmentation task.

\begin{algorithm*}[!t]
\caption{Search results of the best model for each image.}\label{alg:ap_search}
\begin{algorithmic}[1]
\Require{{$ $}
\begin{enumerate}
    \item Collection  of instance segmentation results $ins[N \times K]$ of $K$ models on $N$ images. 
    \item List of ground truths $gt[N]$ of $N$ images. 
\end{enumerate}
}
\Ensure{{$ $}
\begin{enumerate}
    \item Prediction list $pred[N]$ of $N$ images. Each element of $ins$, $gt$, $pred$ contains all segmented results of the corresponding image. %
    \item List of best models $mod[N]$ corresponding to $N$ images.
\end{enumerate}
}
\State $pred \gets \emptyset$ \Comment{Initialize empty prediction list}
\For{$i \gets 1$ to $N$} \Comment{Iterate over images}
\State {$\hat{k} \gets \arg\!\max\limits_{k \in K}($ \textbf{AP} $(gt[1:i], pred \cup ins[i, k]))$} \Comment{Select model (``best model") that achieves highest AP}
\State {$ pred \gets pred \cup ins[i, \hat{k}]$} \Comment{Append segmentation result from best model to prediction list}
\State {$ mod \gets mod \cup \hat{k}$} \Comment{Append ID of best model}
\EndFor
\State \textbf{return} $pred$, $mod$
\end{algorithmic}
\end{algorithm*}

\textbf{Instance Mask Size.} 
The mask size of an instance is defined as the number of pixels in the mask compared with the number in the image. As shown in Figure \ref{fig:ins-size}, the CAMO++ dataset contains a large number of small and medium instances. Small instances (smaller than 0.1) comprise $69.6\%$ of the total number, and medium instances (size from 0.1 to 0.3) comprise $23.8\%$. In the COD dataset, small instances comprise $14.3\%$ of the total number, and medium instances comprise $64.5\%$. Moreover, the CAMO++ dataset contains a fair number of tiny instances, which makes our dataset even more challenging for the camouflaged instance segmentation task. In addition, instance the bounding box distributions in Figure~\ref{fig:bbox-distribution} show that the CAMO++ dataset has a broader range of bounding box sizes than previous datasets.

\textbf{Center Bias.} 
Figure \ref{fig:center_bias} depicts the distributions of object centers in normalized image coordinates over all images in the camouflage datasets. Camouflage instances are biased toward the center of the images in all datasets. This can be explained by the observation that camouflage images are usually cropped for camouflaged instances nearly at the center for people to identify those concealed instances, even for videos (\ie, MoCA dataset~\cite{lamdouar2020betrayed}). Unlike previous datasets, instances in the CAMO++ dataset are localized over the entire image.

The newly constructed CAMO++ dataset inherits challenging attributes from our preliminary CAMO dataset~\cite{ltnghia-CVIU2019}, such as \textit{object appearance, background clutter, shape complexity, object occlusion, and distraction}. A more detailed description is available elsewhere~\cite{ltnghia-CVIU2019}.


\begin{figure}[!t]
    \centering
    \includegraphics[width=\linewidth]{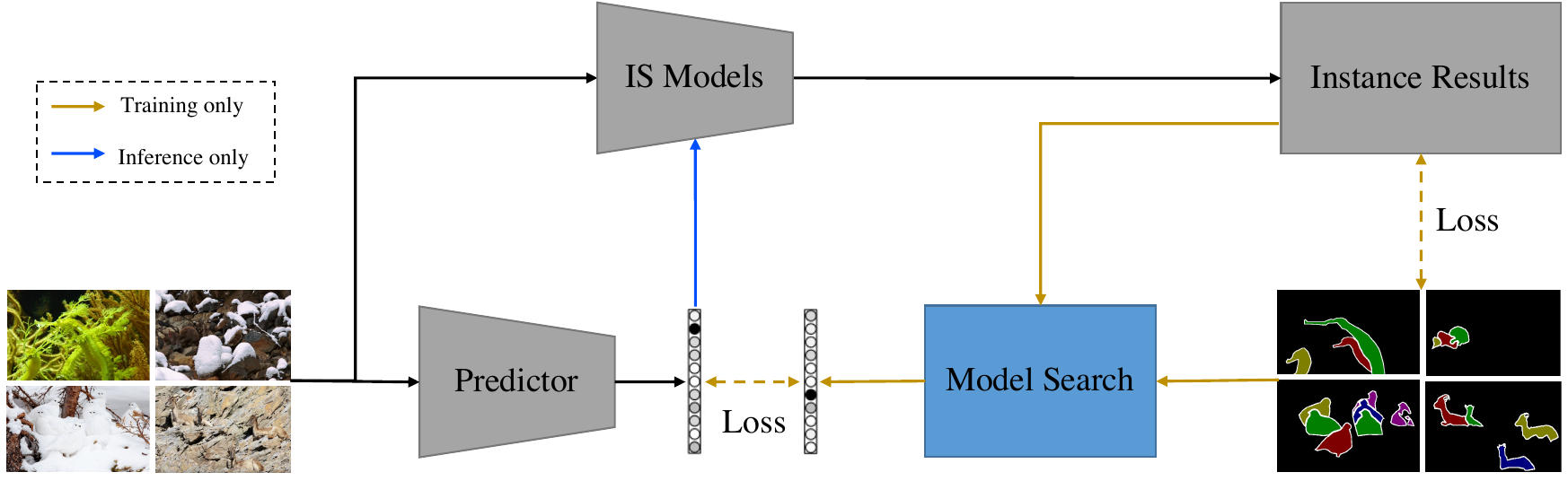}
    \caption{Workflow of camouflage fusion learning (CFL) framework for camouflaged instance segmentation (best viewed online in color). “IS” stands for instance segmentation.}
    \label{fig:multimodal_learning}
\end{figure}

\section{Camouflage Fusion Learning}
\label{section:method}

\subsection{Proposed Framework}
\subsubsection{Overview}

General instance segmentation methods~\cite{Kaiming-ICCV2017, Huang-CVPR2019} can be applied to camouflaged instance segmentation by fine-tuning models on camouflage datasets. However, they are imperfect in the sense that each method may have advantages in specific contexts but disadvantages in others. To utilize the strength of each instance segmentation method, we propose a simple yet efficient CFL framework to fuse various models by learning image contexts.

Figure~\ref{fig:multimodal_learning} depicts our proposed CFL framework for camouflaged instance segmentation. We first train instance segmentation methods on our CAMO++ dataset independently (see Section~\ref{section:compared_methods}). We next generate results for all instance segmentation models and then search for results corresponding to the best model for each image (Algorithm~\ref{alg:ap_search} is our search algorithm). We later train a model predictor to predict the best instance segmentation model for each image.

\subsubsection{Model Search}

Algorithm \ref{alg:ap_search} leverages the insight from a greedy algorithm and iterates over the images to update a ``waiting list." At each iteration, segmentation results of models for an image is added to the waiting list, evaluation is performed to choose the best model, and the corresponding segmentation results are used as pseudo labels to train a model predictor. 

In particular, when working with the $i^{th}$ image, a ``prediction list" contains the segmentation masks for the $1^{st}$ to the $(i-1)^{th}$ image. The waiting list is the union of the prediction list and temporary segmentation mask for the $i^{th}$ image. The average precision (AP)~\cite{Lin-ECCV2014} between the ground truths of the whole training set and the waiting list are then evaluated. If the $k^{th}$ segmentation model is assumed to provide the best AP value, this image will get pseudo label $k$ for the training model predictor afterward. Additionally, the segmentation mask of the $k^{th}$ model for the $i^{th}$ image is appended to the prediction list.

\subsubsection{Objective Function}

Given an image $x$ with corresponding instance ground truth $y$, model predictor $f$, and $M$ instance segmentation models $g$, our loss function comprises two parts:
\begin{equation}
\mathcal{L} = \mathcal{L}_{segm} + \mathcal{L}_{pred},
\end{equation}
where $\mathcal{L}_{segm}$ is instance segmentation loss and $\mathcal{L}_{pred}$ is model prediction loss. The segmentation losses of instance segmentation models $g$ are calculated in accordance with the authors' released source code. Readers could refer to respective works to find more details of the segmentation loss functions.

\begin{equation}
\mathcal{L}_{segm}(x) = \sum_{g =1}^{M} c_g(x) \times \mathcal{L}_{ins}^g(g(x),y),
\end{equation}

where $c$ is a vector in which the best model $i$ selected using Algorithm~\ref{alg:ap_search} is indicated by $c_i = 1$ and the other models are indicated by $0$. The cross entropy loss for multinomial logistic regression~\cite{Katarzyna-TFML2017, Jurafsky-2009, Alexey-ICLR2021} is used as the model prediction loss:
\begin{equation}
\mathcal{L}_{pred}(x) = -c(x) \cdot log(f(x)).
\end{equation}

\subsubsection{Implementation and Training Strategy}

We employed Vision Transformer (ViT-Base16)~\cite{Alexey-ICLR2021} as the model predictor. Five well-known instance segmentation methods: Mask RCNN~\cite{Kaiming-ICCV2017}, Cascade Mask RCNN~\cite{Zhaowei-CVPR2018}, MS RCNN~\cite{Huang-CVPR2019}, RetinaMask~\cite{Fu-2019}, and CenterMask~\cite{Lee-CVPR2020} were empirically chosen for our CFL framework. The framework was developed on PyTorch while the individual models came from publicly available source codes provided by the respective authors.

Our CFL framework was trained in two stages. In the first stage, instance segmentation models were trained independently with their corresponding losses $\mathcal{L}_{ins}$ (see Section~\ref{section:compared_methods}). In the second stage, the instance segmentation models were frozen, the search algorithm was run to select the best ones, and the model predictor was trained.

The predictor training process was conducted by fine-tuning the ImageNet pre-trained model on our CAMO++ dataset. We proposed five-fold stratified sampling for the training strategy. In particular, we randomly split the training data into a training set (4 folds) and a validation set (1 fold) to avoid overfitting. We also applied simple augmentations: resizing, cropping, translation, rotation, and flipping. The ViT model predictor was trained for 100 epochs with a batch size of 16, a base learning rate of 0.0008, a warmup of 1000 steps, and cosine learning rate decay. We used AdamW optimization with a weight decay of $0.001$ and momentum $\beta_1=0.9$ and $\beta_2=0.999$. 

We trained the models on PCs with 64-GB RAM and Tesla P100 GPUs. The training code and trained model will be published upon acceptance of this paper.


\begin{table*}[t!]
\caption{Results of setting 1: Camouflaged instances are not always present. Best results for each type of backbone are shown in \textcolor{ForestGreen}{\textbf{green}}, \textcolor{red}{\textbf{red}}, and \textcolor{blue}{\textbf{blue}}, respectively. }
\resizebox{1\textwidth}{!}{
\begin{tabular}{l|l|ccc|ccc|ccc|ccc}
\toprule
\multirow{2}{*}{\textbf{Method}} & \multirow{2}{*}{\textbf{Backbone}} &
  \multicolumn{3}{c|}{\textbf{Average Precision (AP)}} &
  \multicolumn{3}{c|}{\textbf{AP Across Scales}} &
  \multicolumn{3}{c|}{\textbf{Average Recall (AR)}} &
  \multicolumn{3}{c}{\textbf{AR Across Scales}} \\
\cmidrule{3-14}
 &
 &
  \textbf{AP} &
  \textbf{AP$_{50}$} &
  \textbf{AP$_{75}$} &
  \textbf{AP$_{S}$} &
  \textbf{AP$_{M}$} &
  \textbf{AP$_{L}$} &
  \textbf{AR$_{1}$} &
  \textbf{AR$_{10}$} &
  \textbf{AR$_{100}$} &
  \textbf{AR$_{S}$} &
  \textbf{AR$_{M}$} &
  \textbf{AR$_{L}$} \\
\midrule
Mask RCNN~\cite{Kaiming-ICCV2017} & ResNet50-FPN & 16.1 & 36.4 & 12.9 & 4.0 & 19.0 & 34.1 & 14.7 & 22.3 & 24.3 & 6.4 & 28.6 & 46.0 \\
Mask RCNN~\cite{Kaiming-ICCV2017} & ResNet101-FPN & 17.1 & 37.6 & 14.5 & 4.7 & 20.5 & 35.3 & 15.5 & 23.1 & 25.3 & 6.8 & 31.0 & 47.1 \\
Mask RCNN~\cite{Kaiming-ICCV2017} & ResNeXt101-FPN & 19.9 & 42.2 & 16.7 & 6.8 & 22.5 & 39.7 & 18.5 & 25.2 & 27.0 & 8.4 & 31.4 & 50.2 \\
\midrule
Cascade Mask RCNN~\cite{Zhaowei-CVPR2018} & ResNet50-FPN & 16.9 & 36.6 & 13.9 & 3.6 & 20.0 & 35.8 & 15.5 & 23.5 & 25.6 & 5.3 & 30.5 & 48.3 \\
Cascade Mask RCNN~\cite{Zhaowei-CVPR2018} & ResNet101-FPN & 17.6 & 37.9 & 15.0 & 4.7 & 21.6 & 37.0 & 15.9 & 24.3 & 26.7 & 8.8 & 31.6 & 50.0 \\
Cascade Mask RCNN~\cite{Zhaowei-CVPR2018} & ResNeXt101-FPN & 21.5 & 44.0 & 19.7 & 6.3 & 26.1 & 40.3 & 18.6 & 28.8 & 32.0 & 11.9 & 39.2 & 55.2 \\
\midrule
MS RCNN~\cite{Huang-CVPR2019} & ResNet50-FPN & 19.1 & 38.1 & \textcolor{ForestGreen}{\textbf{18.0}} & 8.3 & 23.2 & 35.4 & 15.5 & 23.0 & 25.9 & 12.1 & 31.3 & 44.9 \\
MS RCNN~\cite{Huang-CVPR2019} & ResNet101-FPN & 20.0 & 38.4 & 18.9 & 8.6 & 24.6 & 37.2 & 16.2 & 24.1 & 26.9 & 12.5 & 33.3 & 47.2 \\
MS RCNN~\cite{Huang-CVPR2019} & ResNeXt101-FPN & 22.9 & 44.2 & 21.8 & 11.8 & 27.8 & 39.9 & 18.0 & 26.9 & 30.5 & 14.6 & 37.1 & 50.2 \\
\midrule
RetinaMask~\cite{Fu-2019} & ResNet50-FPN & 17.4 & 37.5 & 14.5 & 8.8 & 21.8 & 32.9 & 15.1 & 23.8 & 27.7 & 12.5 & 33.6 & 46.3 \\
RetinaMask~\cite{Fu-2019} & ResNet101-FPN & 18.0 & 38.2 & 15.7 & 10.3 & 22.2 & 33.9 & 16.1 & 24.9 & 28.5 & \textcolor{red}{\textbf{18.1}} & 35.6 & 47.8 \\
RetinaMask~\cite{Fu-2019} & ResNeXt101-FPN & 20.0 & 41.2 & 17.6 & \textcolor{blue}{\textbf{12.4}} & 24.9 & 36.9 & 17.8 & 26.6 & 30.4 & \textcolor{blue}{\textbf{16.6}} & 38.0 & 50.5 \\
\midrule
YOLACT~\cite{Bolya-ICCV2019} & ResNet50-FPN & 14.7 & 31.4 & 12.9 & 5.1 & 19.8 & 32.8 & 14.7 & 20.9 & 22.8 & 11.5 & 29.2 & 43.9 \\
YOLACT~\cite{Bolya-ICCV2019} & ResNet101-FPN & 15.9 & 32.4 & 14.2 & 8.6 & 19.6 & 36.1 & 16.1 & 22.2 & 23.6 & 13.0 & 28.1 & 47.1 \\
\midrule
CenterMask~\cite{Lee-CVPR2020} &     ResNet50-FPN & 14.9 & 34.9 & 11.1 & 9.5 & 18.6 & 27.5 & 13.3 & 22.9 & 26.6 & 14.5 & 34.8 & 43.2 \\
CenterMask~\cite{Lee-CVPR2020} & ResNet101-FPN & 15.9 & 35.7 & 12.8 & 8.8 & 19.4 & 29.9 & 14.8 & 23.5 & 26.1 & 14.5 & 33.6 & 44.5 \\
CenterMask~\cite{Lee-CVPR2020} & ResNeXt101-FPN & 19.9 & 42.4 & 17.5 & 9.0 & 24.4 & 33.9 & 16.9 & 26.8 & 30.3 & 15.2 & 38.5 & 47.7 \\
\midrule
SOLO~\cite{Wang-ECCV2020} & ResNet50-FPN & 14.8 & 29.7 & 13.6 & 6.3 & 20.6 & 31.6 & 13.9 & 21.4 & 23.8 & 8.8 & 31.0 & 45.9 \\
SOLO~\cite{Wang-ECCV2020} & ResNet101-FPN & 18.3 & 36.2 & 16.5 & 5.9 & 22.5 & 35.0 & 17.3 & 25.5 & 27.4 & 12.2 & 33.5 & 49.5 \\
SOLO~\cite{Wang-ECCV2020} & ResNeXt101-FPN & 19.3 & 37.3 & 18.2 & 9.3 & 21.1 & 37.5 & 18.7 & 26.8 & 28.6 & 12.0 & 33.2 & 52.7 \\
\midrule
BlendMask~\cite{Chen-CVPR2020} & ResNet50-FPN & 18.2 & 38.6 & 15.4 & \textcolor{ForestGreen}{\textbf{9.8}} & \textcolor{ForestGreen}{\textbf{23.7}} & 34.1 & 15.9 & 24.7 & \textcolor{ForestGreen}{\textbf{28.6}} & \textcolor{ForestGreen}{\textbf{15.3}} & \textcolor{ForestGreen}{\textbf{35.7}} & 47.7 \\
BlendMask~\cite{Chen-CVPR2020} & ResNet101-FPN & 20.3 & 40.6 & 18.2 & \textcolor{red}{\textbf{12.1}} & \textcolor{red}{\textbf{26.5}} & 38.1 & 17.3 & 25.6 & 28.9 & 14.9 & \textcolor{red}{\textbf{36.7}} & 50.9 \\
\midrule
\rowcolor{lightgray} Our CFL framework & ResNet50-FPN & \textcolor{ForestGreen}{\textbf{19.2}} & \textcolor{ForestGreen}{\textbf{39.0}} & 16.5 & 3.1 & 21.8 & \textcolor{ForestGreen}{\textbf{40.8}} & \textcolor{ForestGreen}{\textbf{17.5}} & \textcolor{ForestGreen}{\textbf{25.4}} & 27.9 & 7.9 & 34.2 & \textcolor{ForestGreen}{\textbf{50.0}} \\
\rowcolor{lightgray} Our CFL framework & ResNet101-FPN & \textcolor{red}{\textbf{21.9}} & \textcolor{red}{\textbf{41.9}} & \textcolor{red}{\textbf{21.2}} & 7.3 & 25.5 & \textcolor{red}{\textbf{43.6}} & \textcolor{red}{\textbf{18.5}} & \textcolor{red}{\textbf{26.9}} & \textcolor{red}{\textbf{29.2}} & 12.2 & 36.1 & \textcolor{red}{\textbf{52.0}} \\
\rowcolor{lightgray} Our CFL framework & ResNeXt101-FPN & \textcolor{blue}{\textbf{25.1}} & \textcolor{blue}{\textbf{47.2}} & \textcolor{blue}{\textbf{24.1}} & 5.1 & \textcolor{blue}{\textbf{29.2}} & \textcolor{blue}{\textbf{47.0}} & \textcolor{blue}{\textbf{21.0}} & \textcolor{blue}{\textbf{29.8}} & \textcolor{blue}{\textbf{33.1}} & 11.8 & \textcolor{blue}{\textbf{40.5}} & \textcolor{blue}{\textbf{55.9}} \\
\bottomrule
\end{tabular}
}
\label{table:setting_1}
\end{table*}

\section{Benchmark Suite}
\label{section:benchmark}

\subsection{Baseline Methods}
\label{section:compared_methods}

\begin{figure}[!t]
    \centering
        \includegraphics[width=1\linewidth]{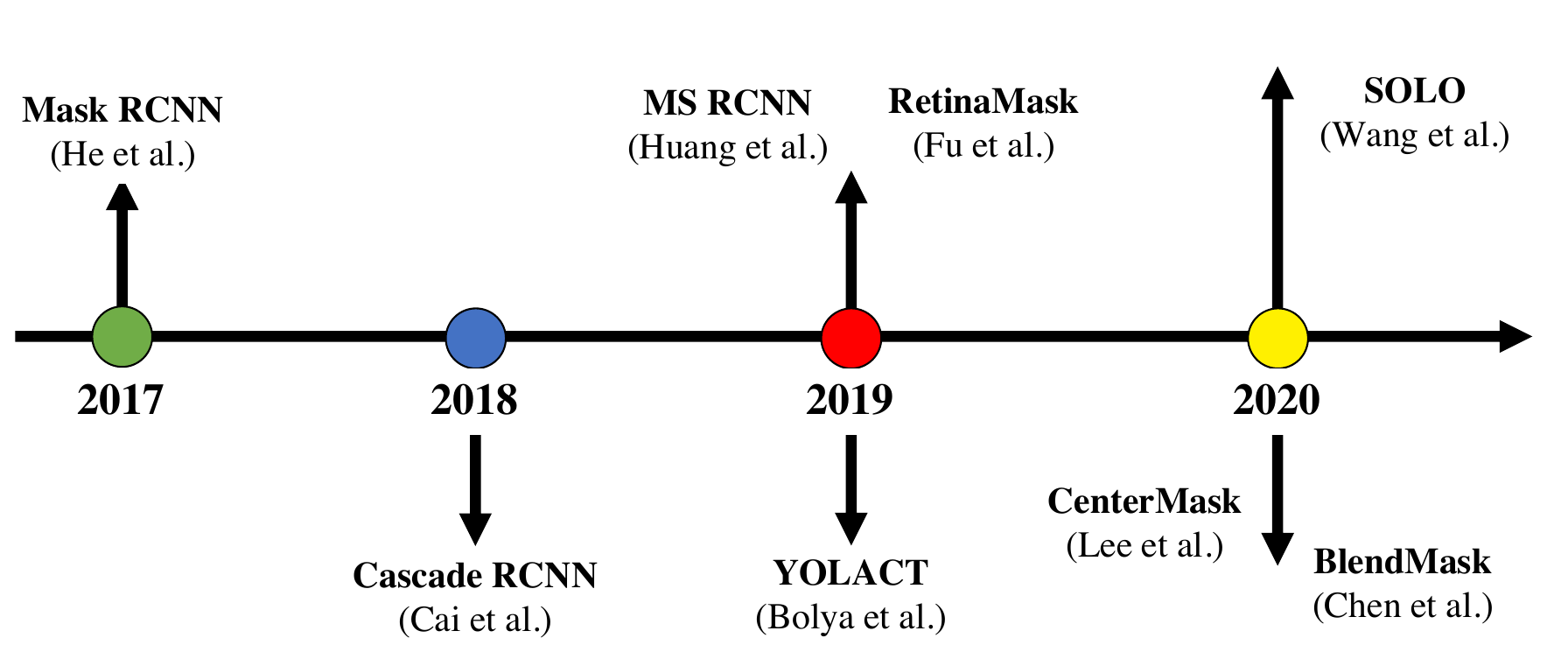}
    \caption{Timeline of benchmarked instance segmentation methods.}
    \label{fig:timeline}
\end{figure}

In addition to constructing our large-scale CAMO++ dataset, we conducted an intensive benchmark for camouflaged instance segmentation. To this end, we trained and tested eight state-of-the-art instance segmentation methods (\ie, Mask RCNN~\cite{Kaiming-ICCV2017}, Cascade Mask RCNN~\cite{Zhaowei-CVPR2018}, MS RCNN~\cite{Huang-CVPR2019}, RetinaMask~\cite{Fu-2019}, YOLACT~\cite{Bolya-ICCV2019},  
CenterMask~\cite{Lee-CVPR2020},  
SOLO~\cite{Wang-ECCV2020}, 
and BlendMask~\cite{Chen-CVPR2020}, 
) on the subsets described in Section~\ref{section:dataset_splits}. Figure~\ref{fig:timeline} shows the development timeline of the benchmarked methods. We used three different backbones (ResNet50-FPN~\cite{He-CVPR2016}, ResNet101-FPN~\cite{He-CVPR2016}, and ResNeXt101-FPN~\cite{Xie-CVPR2017}) for each instance segmentation method. For YOLACT and BlendMask, we used only the ResNet50-FPN and ResNet101-FPN backbones because these methods have not yet been implemented on the ResNeXt101-FPN backbone~\cite{Bolya-ICCV2019, Chen-CVPR2020}.

We trained the models on PCs with 64-GB RAM and Tesla P100 GPUs. The methods were fine-tuned using the MS-COCO pre-trained models and default public configurations provided by the respective authors.

\subsection{Evaluation Measures}

The results given here follow standard COCO-style average precision (AP) metrics: AP (averaged over IoU thresholds from $50\%$ to $95\%$), $AP_{50}$ (AP for IoU threshold 50\%), and $AP_{75}$ (AP for IoU threshold 75\%)~\cite{Lin-ECCV2014}. We also evaluated the results using AP at different scales ($AP_{S}$, $AP_{M}$, and $AP_{L}$), where S, M, and L represent small (area of less than 32 $\times$ 32 pixels), medium (area of 32 $\times$ 32 to 96 $\times$ 96 pixels), and large objects (area of above 96 $\times$ 96 pixels), respectively.

The results also include average recall (AR) metrics: $AR_{1}$, $AR_{10}$, and $AR_{100}$ (AR for given number of results per image)~\cite{Lin-ECCV2014}. Similar to the AP evaluation, we evaluated the AR results at different scales ($AR_{S}$, $AR_{M}$, and $AR_{L}$).

\subsection{Benchmarking Results}

We evaluated the instance segmentation methods on two different settings:
\begin{itemize}
    \item Setting 1: simulates real world (\ie, in-the-wild or unrestricted images), where camouflaged instances are not always present in images. We trained and tested the models on all images. 
    \item Setting 2: camouflaged instances are assumed to be present in every image. We trained and tested the models on only images containing camouflaged instances.
\end{itemize}

\begin{table*}[t!]
\caption{Results of setting 2: Camouflaged instances are present in every image. Best results for each type of backbone are shown in \textcolor{ForestGreen}{\textbf{green}}, \textcolor{red}{\textbf{red}}, and \textcolor{blue}{\textbf{blue}}, respectively.}
\resizebox{1\textwidth}{!}{
\begin{tabular}{l|l|ccc|ccc|ccc|ccc}
\toprule
\multirow{2}{*}{\textbf{Method}} & \multirow{2}{*}{\textbf{Backbone}} & 
  \multicolumn{3}{c|}{\textbf{Average Precision (AP)}} &
  \multicolumn{3}{c|}{\textbf{AP Across Scales}} &
  \multicolumn{3}{c|}{\textbf{Average Recall (AR)}} &
  \multicolumn{3}{c}{\textbf{AR Across Scales}} \\
\cmidrule{3-14}
 &
 &
  \textbf{AP} &
  \textbf{AP$_{50}$} &
  \textbf{AP$_{75}$} &
  \textbf{AP$_{S}$} &
  \textbf{AP$_{M}$} &
  \textbf{AP$_{L}$} &
  \textbf{AR$_{1}$} &
  \textbf{AR$_{10}$} &
  \textbf{AR$_{100}$} &
  \textbf{AR$_{S}$} &
  \textbf{AR$_{M}$} &
  \textbf{AR$_{L}$} \\
\midrule
Mask RCNN~\cite{Kaiming-ICCV2017} & ResNet50-FPN & 24.4 & 52.5 & 21.1 & 6.5 & 18.7 & 26.3 & 26.7 & 33.1 & 33.1 & 6.7 & 25.1 & 35.7 \\
Mask RCNN~\cite{Kaiming-ICCV2017} & ResNet101-FPN & 23.7 & 51.5 & 19.6 & 7.0 & 16.2 & 26.0 & 26.3 & 32.0 & 32.1 & 7.8 & 23.8 & 34.7 \\
Mask RCNN~\cite{Kaiming-ICCV2017} & ResNeXt101-FPN & 31.2 & 62.7 & 29.4 & 8.6 & 22.3 & 34.1 & 32.8 & 38.6 & 38.7 & 11.7 & 28.1 & 42.0 \\
\midrule
Cascade Mask RCNN~\cite{Zhaowei-CVPR2018} & ResNet50-FPN & 25.1 & 51.8 & 21.9 & 4.8 & 17.4 & 27.7 & 27.0 & 34.4 & 34.5 & 4.4 & 24.7 & 37.7 \\
Cascade Mask RCNN~\cite{Zhaowei-CVPR2018} & ResNet101-FPN & 26.7 & 53.8 & 23.6 & 11.4 & 18.9 & 29.2 & 28.8 & 36.5 & 36.5 & 12.2 & 25.8 & 39.8 \\
Cascade Mask RCNN~\cite{Zhaowei-CVPR2018} & ResNeXt101-FPN & 33.6 & 66.0 & 30.7 & 11.7 & 25.0 & 36.5 & 34.0 & 44.6 & 45.0 & 13.9 & 35.0 & 48.2 \\
\midrule
MS RCNN~\cite{Huang-CVPR2019} & ResNet50-FPN & 27.0 & 54.3 & 24.5 & 11.7 & 19.6 & 29.3 & 27.5 & 33.4 & 3.5 & 1.33 & 26.1 & 35.8 \\
MS RCNN~\cite{Huang-CVPR2019} & ResNet101-FPN & 28.1 & 55.6 & 26.2 & 9.3 & 21.7 & 30.2 & 29.0& 35.0 & 35.4 & 10.0 & 27.7 & 37.8 \\
MS RCNN~\cite{Huang-CVPR2019} & ResNeXt101-FPN & 31.0 & 60.4 & 29.2 & 8.0 & 24.2 & 33.2 & 32.4 & 38.4 & 38.6 & 10.0 & 29.8 & 41.5 \\
\midrule
RetinaMask~\cite{Fu-2019} & ResNet50-FPN & 21.5 & 48.2 & 17.0 & \textcolor{ForestGreen}{\textbf{14.4}} & 18.5 & 2.8 & 25.9 & 31.5 & 31.9 & \textcolor{ForestGreen}{\textbf{16.1}} & 25.6 & 33.9 \\
RetinaMask~\cite{Fu-2019} & ResNet101-FPN & 23.5 & 50.2 & 19.5 & 13.9 & 20.3 & 24.9 & 27.4 & 33.7 & 33.9 & 20.0 & 28.6 & 35.6 \\
RetinaMask~\cite{Fu-2019} & ResNeXt101-FPN & 25.8 & 53.1 & 23.3 & 13.3 & 20.5 & 27.8 & 29.9 & 36.6 & 36.8 & 13.9 & 28.7 & 39.3 \\
\midrule
YOLACT~\cite{Bolya-ICCV2019} & ResNet50-FPN & 20.6 & 41.9 & 17.5 & 6.5 & 16.3 & 22.3 & 25.8 & 29.3 & 29.6 & 9.4 & 21.8 & 32.0 \\
YOLACT~\cite{Bolya-ICCV2019} & ResNet101-FPN & 21.7 & 42.1 & 20.0 & 6.2 & 17.2 & 23.5 & 25.9 & 30.1 & 30.1 & 7.2 & 22.8 & 32.4 \\
\midrule
CenterMask~\cite{Lee-CVPR2020} & ResNet50-FPN & 19.7 & 47.1 & 13.7 & 6.7 & 13.8 & 21.4 & 22.9 & 30.7 & 30.9 & 11.0 & 26.8 & 32.3 \\
CenterMask~\cite{Lee-CVPR2020} & ResNet101-FPN & 23.1 & 53.3 & 17.1 & 9.3 & 16.3 & 25.2 & 26.8 & 34.9 & 35.2 & \textcolor{red}{\textbf{26.1}} & 29.0 & 37.0 \\
CenterMask~\cite{Lee-CVPR2020} & ResNeXt101-FPN & 26.6 & 57.4 & 23.2 & 7.7 & 17.0 & 29.6 & 29.8 & 37.1 & 37.3 & 10.0 & 28.7 & 40.0 \\
\midrule
SOLO~\cite{Wang-ECCV2020} & ResNet50-FPN & 22.6 & 46.9 & 19.3 & 6.0 & 8.9 & 26.5 & 26.2 & 28.9 & 29.0 & 6.1 & 13.2 & 33.6 \\
SOLO~\cite{Wang-ECCV2020} & ResNet101-FPN & 23.8 & 48.5 & 21.3 & 6.9 & 12.0 & 27.3 & 28.4 & 30.9 & 31.0 & 7.8 & 15.9 & 35.4 \\
SOLO~\cite{Wang-ECCV2020} & ResNeXt101-FPN & 24.3 & 48.5 & 22.1 & 5.8 & 9.0 & 28.7 & 28.9 & 31.2 & 31.3 & 8.3 & 13.1 & 36.5 \\
\midrule
BlendMask~\cite{Chen-CVPR2020} & ResNet50-FPN & 21.7 & 47.4 & 18.3 & 10.3 & 17.0 & 23.4 & 26.1 & 31.4 & 31.5 & 14.4 & 24.4 & 33.6 \\
BlendMask~\cite{Chen-CVPR2020} & ResNet101-FPN & 25.5 & 53.5 & 20.6 & \textcolor{red}{\textbf{18.8}} & 20.2 & 27.3 & 30.2 & 35.5 & 35.6 & 22.2 & 26.9 & 38.1 \\
\midrule
\rowcolor{lightgray} Our CFL framework & ResNet50-FPN & \textcolor{ForestGreen}{\textbf{35.2}} & \textcolor{ForestGreen}{\textbf{66.3}} & \textcolor{ForestGreen}{\textbf{32.6}} & 6.3 & \textcolor{ForestGreen}{\textbf{25.9}} & \textcolor{ForestGreen}{\textbf{38.2}} & \textcolor{ForestGreen}{\textbf{32.5}} & \textcolor{ForestGreen}{\textbf{39.6}} & \textcolor{ForestGreen}{\textbf{39.8}} & 8.3 & \textcolor{ForestGreen}{\textbf{31.0}} & \textcolor{ForestGreen}{\textbf{42.7}} \\
\rowcolor{lightgray} Our CFL framework & ResNet101-FPN & \textcolor{red}{\textbf{36.9}} & \textcolor{red}{\textbf{68.2}} & \textcolor{red}{\textbf{34.7}} & 15.4 & \textcolor{red}{\textbf{27.0}} & \textcolor{red}{\textbf{39.9}} & \textcolor{red}{\textbf{34.1}} & \textcolor{red}{\textbf{42.2}} & \textcolor{red}{\textbf{42.7}} & 22.8 & \textcolor{red}{\textbf{33.6}} & \textcolor{red}{\textbf{45.4}} \\
\rowcolor{lightgray} Our CFL framework & ResNeXt101-FPN & \textcolor{blue}{\textbf{42.8}} & \textcolor{blue}{\textbf{75.9}} & \textcolor{blue}{\textbf{44.6}} & \textcolor{blue}{\textbf{17.2}} & \textcolor{blue}{\textbf{32.5}} & \textcolor{blue}{\textbf{46.1}} & \textcolor{blue}{\textbf{38.8}} & \textcolor{blue}{\textbf{47.9}} & \textcolor{blue}{\textbf{48.3}} & \textcolor{blue}{\textbf{21.7}} & \textcolor{blue}{\textbf{37.8}} & \textcolor{blue}{\textbf{51.5}} \\
\bottomrule
\end{tabular}
\label{table:setting_2}
}
\end{table*}

\subsubsection{\textbf{Setting 1 (Camouflaged Instances Are Not Always Present)}}

\begin{figure*}[!t]
	\centering
	\includegraphics[width=1\linewidth]{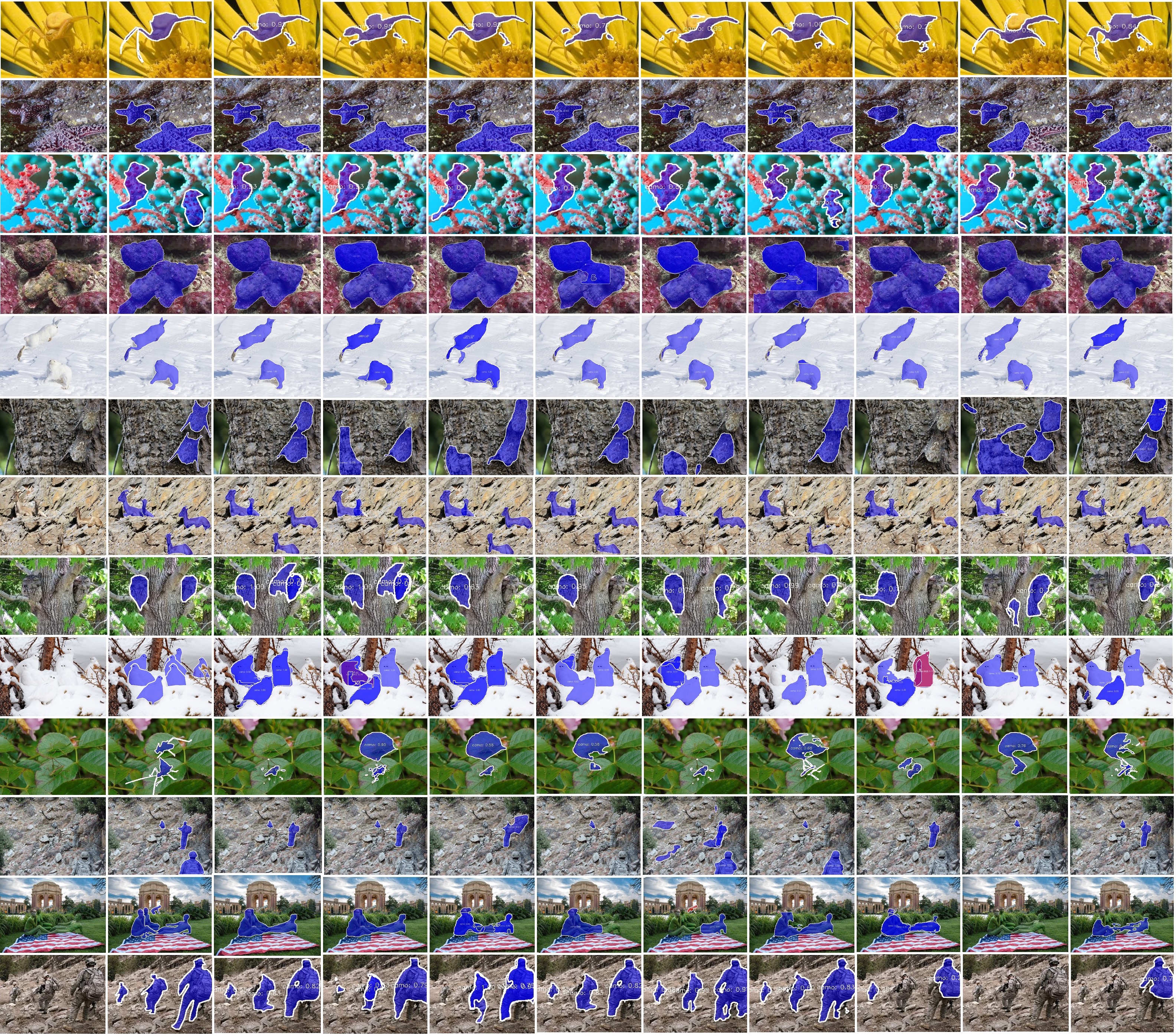}
	\caption{Comparison of results of eight state-of-the-art instance segmentation methods. From left to right: original image is followed by overlaid ground truth and results of our CFL framework, Mask RCNN~\cite{Kaiming-ICCV2017}, Cascade Mask RCNN~\cite{Zhaowei-CVPR2018}, MS RCNN~\cite{Huang-CVPR2019}, RetinaMask~\cite{Fu-2019}, YOLACT~\cite{Bolya-ICCV2019}, CenterMask~\cite{Lee-CVPR2020}, SOLO~\cite{Wang-ECCV2020}, and BlendMask~\cite{Chen-CVPR2020}. Camouflaged instances are shown in blue, and non-camouflaged instances are shown in red. Best viewed in color with zoom.}
	\label{fig:visualized_results}
\end{figure*}

\begin{figure*}[!t]
	\centering
	\includegraphics[width=1\linewidth]{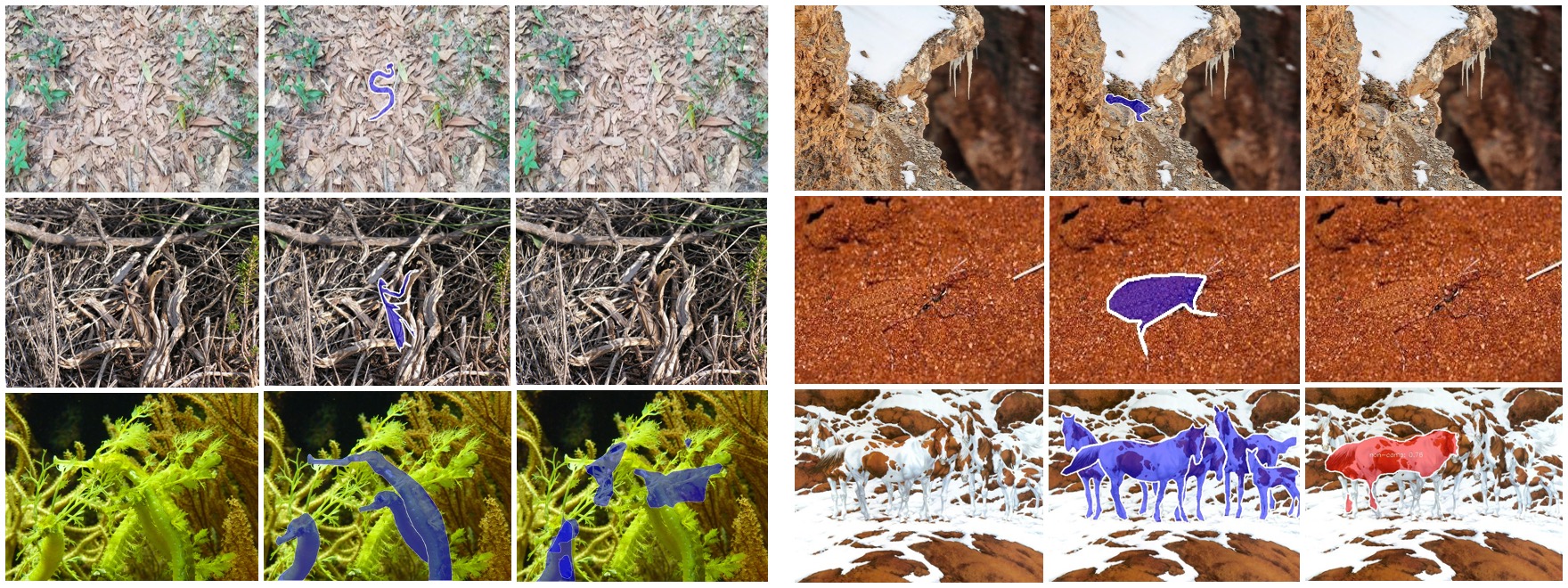}
	\caption{Failure cases in camouflaged instance segmentation. From left to right: original image is followed by overlaid ground truth and results of our CFL framework. Camouflaged instances are shown in blue, and non-camouflaged instances are shown in red. From top to bottom are examples of tiny instances, extreme resemblance to the background, and occluded and overlapping camouflaged instances, which are immensely challenging even for human detection. Best viewed in color with zoom.}
	\label{fig:failure}
\end{figure*}

Table~\ref{table:setting_1} details the performance of the tested methods for the first experimental setting: segment multiple camouflaged instances on unrestricted images. As can be seen, better backbones tended to produce better results within the same method. The ResNeXt101-based implementations had the best performance. In terms of AP, the AP$_{50}$ metric had the highest scores whereas the AP$_{75}$ metric had the lowest scores. The AP$_{L}$, AR$_{100}$, and AR$_{L}$ metrics had the highest scores in terms of AP Across Scales, AR, and AR Across Scales, respectively. In general, the benchmarking methods consistently performed across the performance metrics. In other words, the methods performing well for one performance metric tended to perform well for the others. There was no dominant method across all metrics. For example, RetinaMask surpassed BlendMask in terms of \textbf{AP}$_S$ and \textbf{AR}$_S$ whereas MS RCNN surpassed RetinaMask in terms of \textbf{AP}$_{75}$, \textbf{AP}$_M$, and \textbf{AP}$_L$. 

Our fusion method is aimed at leveraging the advantages of different methods in order to produce the best results. Our proposed scene-driven framework selects the best models adaptively for each image by learning its visual deep features. It can thus take advantage of all component models, resulting in superior performance. In particular, our CFL framework achieved the state-of-the-art performance across all metrics. It significantly outperformed the other instance segmentation methods with an AP of 19.2, 21.9, and 25.1 on the ResNet50-FPN, ResNet101-FPN, and ResNeXt101-FPN backbones, respectively. It was also consistently better than the others in terms of AR.

\subsubsection{\textbf{Setting 2 (Camouflaged Instances Are Always Present)}}

Table~\ref{table:setting_2} details the performance of the tested methods for the second experimental setting (in which camouflaged objects are assumed to be present in every image). Again, thanks to multimodal fusion learning, the CFL framework consistently outperformed the others. It achieved the best performance across all metrics. In particular, it achieved the state-of-the-art performance with an AP of 35.2, 36.9, and 42.8 on the ResNet50-FPN, ResNet101-FPN, and ResNeXt101-FPN backbones, respectively.

\begin{table}[t]
\centering
\caption{Dataset generalization evaluation of camouflaged instance segmentation, using Cascade RCNN model and AP metric. Columns indicate generalizability of training dataset, measured as average value of testing datasets (higher is better). Rows indicate difficulty of testing dataset, measured as average value of training datasets (lower is better). Best results are shown in \textbf{bold}.}
\resizebox{1\linewidth}{!}{
\begin{tabular}{l|cc|c}
\toprule
\diagbox{Testing Set}{Training Set} & CAMO++ & COD~\cite{Fan-CVPR2020} & \cellcolor{lightgray} Mean-Difficulty $\Downarrow$ \\
\midrule
CAMO++ & 33.6 & 23.2 & \cellcolor{lightgray} \textbf{28.4} \\
COD~\cite{Fan-CVPR2020} & 27.9 & 31.4 & \cellcolor{lightgray} 29.7 \\
\midrule
\rowcolor{lightgray} Mean-Generalizability $\Uparrow$ & \textbf{30.8} & 27.3 &  \\
\bottomrule
\end{tabular}
}
\label{tab:generalization_cis}
\end{table}

\begin{table*}[t!]
\centering
\caption{Results of extra training using non-camouflage images. Best results for each type of backbone are shown in \textcolor{ForestGreen}{\textbf{green}}, \textcolor{red}{\textbf{red}}, and \textcolor{blue}{\textbf{blue}}, respectively. }
\begin{tabular}{l|l|ccc||cc|cc|cc}
\toprule
\multirow{2}{*}{\textbf{Method}} & \multirow{2}{*}{\textbf{Backbone}} & \multicolumn{3}{c||}{\textbf{Without Extra Training}} & \multicolumn{6}{c}{\textbf{With Extra Training}} \\ \cmidrule{3-11}
 &
 &
  \textbf{AP} &
  \textbf{AP$_{50}$} &
  \textbf{AP$_{75}$} &
  \textbf{AP} & &
  \textbf{AP$_{50}$} & & 
  \textbf{AP$_{75}$} & \\
\midrule
Mask RCNN~\cite{Kaiming-ICCV2017} & ResNet50-FPN & 24.4 & 52.5 & 21.1 &  24.6 & (+0.2) & 53.3 & (+0.8) & 20.8 & (-0.3) \\
Mask RCNN~\cite{Kaiming-ICCV2017} & ResNet101-FPN & 23.7 & 51.5 & 19.6 &  26.1 & (+2.4) & 54.7 & (+3.2) & 23.4 & (+3.8) \\
Mask RCNN~\cite{Kaiming-ICCV2017} & ResNeXt101-FPN & 31.2 & 62.7 & 29.4 &  31.3 & (+0.1) & 63.3 & (+0.6) & 27.7 & (-1.7) \\
\midrule
Cascade Mask RCNN~\cite{Zhaowei-CVPR2018} & ResNet50-FPN & 25.1 & 51.8 & 21.9 &  25.9 & (+0.8) & 53.4 & (+1.6) & 22.4 & (+0.5)  \\
Cascade Mask RCNN~\cite{Zhaowei-CVPR2018} & ResNet101-FPN & 26.7 & 53.8 & 23.6 &  26.9 & (+0.2) & 54.5 & (+0.7) & 24.2 & (+0.6)  \\
Cascade Mask RCNN~\cite{Zhaowei-CVPR2018} & ResNeXt101-FPN & 33.6 & 66.0 & 30.7 & 33.8 & (+0.2) & 64.8 & (-1.2) & 33.0 & (+2.3) \\
\midrule
MS RCNN~\cite{Huang-CVPR2019} & ResNet50-FPN & 27.0 & 54.3 & 24.5 &  26.9 & (-0.1) & 53.2 & (-1.1) & 25.6 & (-1.3) \\
MS RCNN~\cite{Huang-CVPR2019} & ResNet101-FPN & 28.1 & 55.6 & 26.2 &  28.6 & (+0.5) & 54.6 & (-1.0) & 27.7 & (-0.9) \\
MS RCNN~\cite{Huang-CVPR2019} & ResNeXt101-FPN & 31.0 & 60.4 & 29.2  &  32.9 & (+1.9) & 62.4 & (+2.0) & 32.0 & (+2.8) \\
\midrule
RetinaMask~\cite{Fu-2019} & ResNet50-FPN & 21.5 & 48.2 & 17.0 &  23.5 & (+2.0) & 50.9 & (+2.7) & 19.2 & (+2.2) \\
RetinaMask~\cite{Fu-2019} & ResNet101-FPN & 23.5 & 50.2 & 19.5 &  25.4 & (+1.9) & 54.0 & (+3.8) & 21.9 & (+2.4) \\
RetinaMask~\cite{Fu-2019} & ResNeXt101-FPN & 25.8 & 53.1 & 23.3 &  28.1 & (+2.3) & 58.3 & (+5.2) & 24.4 & (+1.1) \\
\midrule
YOLACT~\cite{Bolya-ICCV2019} & ResNet50-FPN & 20.6 & 41.9 & 17.5 &  20.6 & (+0) & 44.3 & (+2.4) & 18.2 & (+0.7) \\
YOLACT~\cite{Bolya-ICCV2019} & ResNet101-FPN & 21.7 & 42.1 & 20.0 &  23.1 & (+1.4) & 46.9 & (+4.8) & 20.7 & (+0.7) \\
\midrule
CenterMask~\cite{Lee-CVPR2020} & ResNet50-FPN & 19.7 & 47.1 & 13.7  &  21.3 & (+1.6) & 50.4 & (+3.3) & 15.7 & (+2.0) \\
CenterMask~\cite{Lee-CVPR2020} & ResNet101-FPN & 23.1 & 53.3 & 17.1  &  24.6 & (+1.5) & 54.6 & (+1.3) & 20.2 & (+3.1) \\
CenterMask~\cite{Lee-CVPR2020} & ResNeXt101-FPN & 26.6 & 57.4 & 23.2  &  29.9 & (+3.3) & 62.7 & (+5.3) & 27.0 & (+3.8) \\
\midrule
SOLO~\cite{Wang-ECCV2020} & ResNet50-FPN & 22.6 & 46.9 & 19.3 &  23.2 & (+0.6) & 46.9 & (+0) & 21.1 & (+1.8) \\
SOLO~\cite{Wang-ECCV2020} & ResNet101-FPN & 23.8 & 48.5 & 21.3 &  30.5 & (+6.7) & 60.0 & (+11.5) & 27.4 & (+6.1) \\
SOLO~\cite{Wang-ECCV2020} & ResNeXt101-FPN & 24.3 & 48.5 & 22.1 &  33.3 & (+9.0) & 62.9 & (+14.4) & 31.8 & (+9.7) \\
\midrule
BlendMask~\cite{Chen-CVPR2020} & ResNet50-FPN & 21.7 & 47.4 & 18.3 &  25.9 & (+4.2) & 55.5 & (+8.1) & 21.6 & (-4.3) \\
BlendMask~\cite{Chen-CVPR2020} & ResNet101-FPN & 25.5 & 53.5 & 20.6  &  28.1 & (+2.6) & 56.5 & (+3.0) & 25.0 & (+4.4) \\
\midrule
\rowcolor{lightgray} Our CFL framework & ResNet50-FPN & \textcolor{ForestGreen}{\textbf{35.2}} & \textcolor{ForestGreen}{\textbf{66.3}} & \textcolor{ForestGreen}{\textbf{32.6}} & \textcolor{ForestGreen}{\textbf{35.7}} & (+0.5) & \textcolor{ForestGreen}{\textbf{65.9}} & (-0.4) & \textcolor{ForestGreen}{\textbf{34.5}} & (+1.9) \\
\rowcolor{lightgray} Our CFL framework & ResNet101-FPN & \textcolor{red}{\textbf{36.9}} & \textcolor{red}{\textbf{68.2}} & \textcolor{red}{\textbf{34.7}} & \textcolor{red}{\textbf{37.6}} & (+0.7) & \textcolor{red}{\textbf{69.2}} & (+1.0) & \textcolor{red}{\textbf{37.8}} & (+3.1) \\
\rowcolor{lightgray} Our CFL framework & ResNeXt101-FPN & \textcolor{blue}{\textbf{42.8}} & \textcolor{blue}{\textbf{75.9}} & \textcolor{blue}{\textbf{44.6}} & \textcolor{blue}{\textbf{43.7}} & (+0.9) & \textcolor{blue}{\textbf{74.8}} & (-1.1) & \textcolor{blue}{\textbf{45.4}} & (+0.8) \\
\bottomrule
\end{tabular}
\label{table:extra_training}
\end{table*}

\subsubsection{\textbf{Qualitative Comparison}} 

Figure \ref{fig:visualized_results} shows a visual comparison of the tested methods on the ResNet50 backbone as it is the most commonly used backbone for segmentation. The CFL framework achieved the best results, and the results are close to the ground truth. The CFL framework was able to segment camouflaged instances with fine details, demonstrating its robustness. It effectively handled a variety of challenging cases, including camouflaged instances with colors and textures similar to those of the background and images with complex shapes and multiple objects.

\subsubsection{\textbf{Failure Cases and Discussion}}
\label{section:failure}

The CFL framework also had failed camouflaged instance segmentations on the CAMO++ dataset, as shown by the examples in Figure~\ref{fig:failure}. It had trouble localizing and segmenting camouflaged instances due to a tiny instances (top row) and extreme resemblance to the background (second row). These cases are immensely challenging even for human detection. There were also failures on occluded or overlapping camouflaged instances (bottom row), resulting in incorrect segmentation (bottom left example) or misclassification between camouflaged instances and non-camouflaged instances (bottom right example).

From the visualization and aforementioned observations along with experimental results, we conclude that even the leading instance segmentation methods in the deep learning era remain limited. They cannot yet effectively segment multiple camouflaged instances on unrestricted images without any assumption (Top-1: $AP \leq 25$ as in Table~\ref{table:setting_1}). Hence, \textit{\textbf{accurate camouflaged instance segmentation of in-the-wild images is still far from being achieved, leaving much room for improvement }}. The results also indicate the challenges posed by our CAMO++ dataset.

\section{Data Analysis}
\label{sec:ablation}

\subsection{Extra Training Using Non-Camouflage Images}

The lack of training data may affect the training of camouflage localization and segmentation systems. Previous camouflage research has used extra images along with augmented data to improve segmentation performance. Fan \etal~\cite{Fan-CVPR2020} combined images from multiple camouflage datasets to train their network. Li \etal~\cite{Aixuan-CVPR2021} recently used the relationship between saliency and camouflage to train a network to jointly segment salient and camouflaged objects.

In this work, we further utilized the information of the non-camouflage images with only general objects to segment camouflaged instances. In particular, we trained instance segmentation methods on a combination of camouflage and non-camouflage images in our CAMO++ dataset.

Table~\ref{table:extra_training} compares the performance of different methods. The addition of the non-camouflaged instances resulted in the fine-tuned models achieving better performance in segmenting the camouflaged instances. This shows that training using additional non-camouflaged instance data helps boost performance. Furthermore, once again, our proposed CFL framework was the top performer. In particular, the performance of the CFL framework with the ResNeXt101-FPN backbone was boosted thanks to the combined data more than the others. It achieved the best performance (AP of 42.8).

\subsection{Dataset Generalization Evaluation}

In addition, we investigates dataset bias through cross-dataset generalization evaluation over the CAMO++ and COD datasets. We used only the camouflage images in both datasets because the COD dataset does not have the ground truths for the non-camouflage images. For a fair comparison, we randomly select 1700 images from the training set and 1000 images from the test set of the COD dataset to obtain the same number of images as in the CAMO++ dataset. We trained Cascade Mask RCNN~\cite{Zhaowei-CVPR2018} with the ResNeXt101-FPN backbone on each dataset using the training configuration described in Section~\ref{section:compared_methods}.

Table \ref{tab:generalization_cis} shows the AP obtained by cross-dataset generalization. Each column illustrates the results of a model that was trained on one dataset and tested on all datasets, indicating the generalizability of the training dataset. Each row shows the performance of a model that was trained on all datasets and tested on a specific dataset, indicating the difficulty of the testing dataset. The results show that our newly constructed CAMO++ dataset is more challenging than the COD dataset. In particular, our training images are unbiased (\eg mean value of 30.8 on the bottom row), which should help boost performance on both the CAMO++ and COD datasets. The results also show that our testing images are the most difficult (\eg mean value of 28.4 in the right column) as they consist of many challenging cases such as tiny, extreme background resemblance, distraction, and occluded and overlapping camouflaged instances (see Section~\ref{section:failure} and Figure~\ref{fig:failure}).


\section{Conclusion and Outlook} \label{section:conclusion}

In our study of the interesting yet challenging problem of camouflaged instance segmentation, we created a large-scale dataset dubbed Camouflaged Object Plus Plus (CAMO++). We also performed an in-depth analysis of CAMO++ to demonstrate its diversity and complexity and developed a camouflage fusion learning framework to further improve the performance of camouflaged instance segmentation. 

We conducted an extensive benchmark for the camouflaged instance segmentation task and evaluated state-of-the-art instance segmentation methods in various experimental settings. We found that using non-camouflaged instances for training boosted the performance of state-of-the-art methods. However, there is still room for improvement, as shown by the cases of failure. The CAMO++ dataset should serve as a valuable benchmark for not only the camouflaged instance segmentation task but also related tasks such as semantic camouflage segmentation and video camouflaged instance segmentation. We expect that our CAMO++ dataset will greatly support research activities in this field.

We intend to explore the effect of various factors on the given problem. For example, the use of contextual information may be helpful in detecting and segmenting camouflaged instances. We also plan to extend our work to dynamic scenes such as those in videos. In particular, we intend to investigate the use of motion information in segmenting camouflaged instances in videos. 

\section*{Acknowledgment}

This work was partially funded by National Science Foundation Grant (NSF\#2025234), Gia Lam Urban Development and Investment Company Limited, Vingroup and supported by Vingroup Innovation Foundation (VINIF) under project code VINIF.2019.DA19. The first author would like to thank JSPS KAKENHI Grants (JP16H06302, JP18H04120, JP21H04907, JP20K23355, JP21K18023), JST CREST Grants (JPMJCR20D3, JPMJCR18A6).

We are grateful to the Software Engineering Lab (University of Science, VNU-HCM) for their support in annotating the CAMO++ dataset. We gratefully acknowledge NVIDIA for their support of the GPUs.

\ifCLASSOPTIONcaptionsoff
  \newpage
\fi

\small{
\bibliography{short_bibtex}

\begin{thebibliography}{10}
\providecommand{\url}[1]{#1}
\csname url@samestyle\endcsname
\providecommand{\newblock}{\relax}
\providecommand{\bibinfo}[2]{#2}
\providecommand{\BIBentrySTDinterwordspacing}{\spaceskip=0pt\relax}
\providecommand{\BIBentryALTinterwordstretchfactor}{4}
\providecommand{\BIBentryALTinterwordspacing}{\spaceskip=\fontdimen2\font plus
\BIBentryALTinterwordstretchfactor\fontdimen3\font minus
  \fontdimen4\font\relax}
\providecommand{\BIBforeignlanguage}[2]{{%
\expandafter\ifx\csname l@#1\endcsname\relax
\typeout{** WARNING: IEEEtran.bst: No hyphenation pattern has been}%
\typeout{** loaded for the language `#1'. Using the pattern for}%
\typeout{** the default language instead.}%
\else
\language=\csname l@#1\endcsname
\fi
#2}}
\providecommand{\BIBdecl}{\relax}
\BIBdecl

\bibitem{ltnghia-CVIU2019}
T.-N. Le, T.~V. Nguyen, Z.~Nie, M.-T. Tran, and A.~Sugimoto, ``Anabranch
  network for camouflaged object segmentation,'' \emph{CVIU}, vol. 184, pp.
  45--56, 2019.

\bibitem{Sujit-ICEECS2013}
S.~Singh, C.~Dhawale, and S.~Misra, ``Survey of object detection methods in
  camouflaged image,'' \emph{IERI Procedia}, vol.~4, pp. 351 -- 357, 2013.

\bibitem{Hemin-ISMICT2019}
H.~A. Qadir, Y.~Shin, J.~Solhusvik, J.~Bergsland, L.~Aabakken, and
  I.~Balasingham, ``Polyp detection and segmentation using mask r-cnn: Does a
  deeper feature extractor cnn always perform better?'' in \emph{ISMICT}, 2019.

\bibitem{Ferhat-MH2020}
F.~Ucar and D.~Korkmaz, ``Covidiagnosis-net: Deep bayes-squeezenet based
  diagnosis of the coronavirus disease 2019 (covid-19) from x-ray images,''
  \emph{Medical Hypotheses}, 2020.

\bibitem{nhhuy-BTAS2019}
H.~H. Nguyen, F.~Fang, J.~Yamagishi, and I.~Echizen, ``Multi-task learning for
  detecting and segmenting manipulated facial images and videos,'' in
  \emph{BTAS}, 2019.

\bibitem{ltnghia-ICCV2021}
T.-N. Le, H.~H. Nguyen, J.~Yamagishi, and I.~Echizen, ``Openforensics:
  Large-scale challenging dataset for multi-face forgery detection and
  segmentation in-the-wild,'' in \emph{ICCV}, 2021.

\bibitem{Kervrann-TIP1995}
C.~Kervrann and F.~Heitz, ``A markov random field model-based approach to
  unsupervised texture segmentation using local and global spatial
  statistics,'' \emph{IEEE TIP}, vol.~4, no.~6, pp. 856--862, 1995.

\bibitem{Boykov-IJCV2006}
Y.~Boykov and G.~Funka-Lea, ``Graph cuts and efficient nd image segmentation,''
  \emph{IJCV}, vol.~70, no.~2, pp. 109--131, 2006.

\bibitem{Li-ICASSP2011}
X.~Li and H.~Sahbi, ``Superpixel-based object class segmentation using
  conditional random fields,'' in \emph{ICASSP}, 2011, pp. 1101--1104.

\bibitem{Sulimowicz-ICIP2018}
L.~Sulimowicz, I.~Ahmad, and A.~Aved, ``Superpixel-enhanced pairwise
  conditional random field for semantic segmentation,'' in \emph{ICIP}, 2018,
  pp. 271--275.

\bibitem{Galun-ICCV2003}
M.~Galun, E.~Sharon, R.~Basri, and A.~Brandt, ``Texture segmentation by
  multiscale aggregation of filter responses and shape elements,'' in
  \emph{ICCV}, Oct 2003, pp. 716--723.

\bibitem{Song-ICMT2010}
L.~Song and W.~Geng, ``A new camouflage texture evaluation method based on
  wssim and nature image features,'' in \emph{International Conference on
  Multimedia Technology}, Oct 2010, pp. 1--4.

\bibitem{Xue-MTA2016}
F.~Xue, C.~Yong, S.~Xu, H.~Dong, Y.~Luo, and W.~Jia, ``Camouflage performance
  analysis and evaluation framework based on features fusion,''
  \emph{Multimedia Tools and Applications}, vol.~75, pp. 4065--4082, 2016.

\bibitem{Pan-MAS2011}
Y.~Pan, Y.~Chen, Q.~Fu, P.~Zhang, and X.~Xu, ``Study on the camouflaged target
  detection method based on 3d convexity,'' \emph{Modern Applied Science},
  vol.~5, no.~4, p. 152, 2011.

\bibitem{Liu-TIP2012}
Z.~Liu, K.~Huang, and T.~Tan, ``Foreground object detection using top-down
  information based on em framework,'' \emph{IEEE TIP}, vol.~21, no.~9, pp.
  4204--4217, Sept 2012.

\bibitem{Sengottuvelan-ICETET2008}
A.~W. P.~Sengottuvelan and A.~Shanmugam, ``Performance of decamouflaging
  through exploratory image analysis,'' in \emph{ICETET}, 2008, pp. 6--10.

\bibitem{Yin-PE2011}
J.~Yin, Y.~Han, W.~Hou, and J.~Li, ``Detection of the mobile object with
  camouflage color under dynamic background based on optical flow,''
  \emph{Procedia Engineering}, vol.~15, pp. 2201 -- 2205, 2011.

\bibitem{Gallego-ICIP2014}
J.~Gallego and P.~Bertolino, ``Foreground object segmentation for moving camera
  sequences based on foreground-background probabilistic models and prior
  probability maps,'' in \emph{ICIP}, Oct 2014, pp. 3312--3316.

\bibitem{Fan-CVPR2020}
D.-P. Fan, G.-P. Ji, G.~Sun, M.-M. Cheng, J.~Shen, and L.~Shao, ``Camouflaged
  object detection,'' in \emph{CVPR}, 2020.

\bibitem{Lamdouar-ACCV2020}
H.~Lamdouar, C.~Yang, W.~Xie, and A.~Zisserman, ``Betrayed by motion:
  Camouflaged object discovery via motion segmentation,'' in \emph{ACCV},
  November 2020.

\bibitem{Jinchao-AAAI2021}
J.~Zhu, X.~Zhang, S.~Zhang, and J.~Liu, ``Inferring camouflage objects by
  texture-aware interactive guidance network,'' in \emph{AAAI}, 2021, pp.
  3599--3607.

\bibitem{ltnghia-AAAI2021}
T.-N. Le, V.~Nguyen, C.~Le, T.-C. Nguyen, M.-T. Tran, and T.~V. Nguyen,
  ``Camoufinder: Finding camouflaged instances in images,'' in \emph{AAAI},
  2021, pp. 16\,071--16\,074.

\bibitem{Bideau-ECCV2016}
E.~L.-M. Pia~Bideau, ``It's moving! a probabilistic model for causal motion
  segmentation in moving camera videos,'' in \emph{ECCV}, 2016.

\bibitem{Skurowski-2018}
P.~Skurowski, H.~Abdulameer, J.~Baszczyk, T.~Depta, A.~Kornacki, and P.~Kozie,
  ``Animal camouflage analysis: Chameleon database,'' \emph{Unpublished
  Manuscript}, 2018.

\bibitem{Jinnan-IEEEAccess2021}
J.~Yan, T.-N. Le, K.-D. Nguyen, M.-T. Tran, T.-T. Do, and T.~V. Nguyen,
  ``Mirrornet: Bio-inspired camouflaged object segmentation,'' \emph{IEEE
  Access}, vol.~9, pp. 43\,290--43\,300, 2021.

\bibitem{Kirillov-CVPR2017}
A.~Kirillov, E.~Levinkov, B.~Andres, B.~Savchynskyy, and C.~Rother,
  ``Instancecut: from edges to instances with multicut,'' in \emph{CVPR},
  vol.~3, 2017, p.~9.

\bibitem{Levinkov-CVPR2017}
E.~Levinkov, J.~Uhrig, S.~Tang, M.~Omran, E.~Insafutdinov, A.~Kirillov,
  C.~Rother, T.~Brox, B.~Schiele, and B.~Andres, ``Joint graph decomposition \&
  node labeling: Problem, algorithms, applications,'' in \emph{CVPR}, 2017.

\bibitem{Liang-TPAMI2017}
X.~Liang, L.~Lin, Y.~Wei, X.~Shen, J.~Yang, and S.~Yan, ``Proposal-free network
  for instance-level semantic object segmentation,'' \emph{IEEE TPAMI}, pp.
  1--1, 2017.

\bibitem{Zhang-CVPR2016}
Z.~Zhang, S.~Fidler, and R.~Urtasun, ``Instance-level segmentation for
  autonomous driving with deep densely connected mrfs,'' in \emph{CVPR}, 2016,
  pp. 669--677.

\bibitem{Konstantin-ICCV2019}
A.~K. Konstantin~Sofiiuk, Olga~Barinova, ``Adaptis: Adaptive instance selection
  network,'' in \emph{ICCV}, 2019.

\bibitem{Dai-CVPR2016}
J.~Dai, K.~He, and J.~Sun, ``Instance-aware semantic segmentation via
  multi-task network cascades,'' in \emph{CVPR}, 2016.

\bibitem{Kaiming-ICCV2017}
K.~He, G.~Gkioxari, P.~Doll{\'a}r, and R.~Girshick, ``Mask r-cnn,'' in
  \emph{ICCV}, 2017, pp. 2980--2988.

\bibitem{Yi-CVPR2017}
Y.~Li, H.~Qi, J.~Dai, X.~Ji, and Y.~Wei, ``Fully convolutional instance-aware
  semantic segmentation,'' in \emph{CVPR}, 2017.

\bibitem{Ren-NeurIPS2015}
S.~Ren, K.~He, R.~Girshick, and J.~Sun, ``Faster r-cnn: Towards real-time
  object detection with region proposal networks,'' in \emph{NeurIPS}, 2015,
  pp. 91--99.

\bibitem{Dai-NeurIPS2016}
J.~Dai, Y.~Li, K.~He, and J.~Sun, ``R-fcn: Object detection via region-based
  fully convolutional networks,'' in \emph{NeurIPS}, 2016, pp. 379--387.

\bibitem{Huang-CVPR2019}
Z.~Huang, L.~Huang, Y.~Gong, C.~Huang, and X.~Wang, ``{Mask Scoring R-CNN},''
  in \emph{CVPR}, 2019.

\bibitem{Zhaowei-CVPR2018}
Z.~Cai and N.~Vasconcelos, ``Cascade r-cnn: Delving into high quality object
  detection,'' in \emph{CVPR}, 2018.

\bibitem{Liu-CVPR2018}
S.~Liu, L.~Qi, H.~Qin, J.~Shi, and J.~Jia, ``Path aggregation network for
  instance segmentation,'' in \emph{CVPR}, 2018.

\bibitem{Zhou-2019}
X.~Zhou, D.~Wang, and P.~Kr{\"a}henb{\"u}hl, ``Objects as points,'' in
  \emph{arXiv preprint arXiv:1904.07850}, 2019.

\bibitem{Tian-ICCV2019}
Z.~Tian, C.~Shen, H.~Chen, and T.~He, ``Fcos: Fully convolutional one-stage
  object detection,'' in \emph{ICCV}, October 2019.

\bibitem{Bolya-ICCV2019}
D.~Bolya, C.~Zhou, F.~Xiao, and Y.~J. Lee, ``Yolact: {Real-time} instance
  segmentation,'' in \emph{ICCV}, 2019.

\bibitem{Chen-CVPR2020}
H.~Chen, K.~Sun, Z.~Tian, C.~Shen, Y.~Huang, and Y.~Yan, ``Blendmask: Top-down
  meets bottom-up for instance segmentation,'' in \emph{CVPR}, 2020.

\bibitem{Lee-CVPR2020}
Y.~Lee and J.~Park, ``Centermask: Real-time anchor-free instance
  segmentation,'' in \emph{CVPR}, 2020.

\bibitem{Xie-CVPR2020}
E.~Xie, P.~Sun, X.~Song, W.~Wang, X.~Liu, D.~Liang, C.~Shen, and P.~Luo,
  ``Polarmask: Single shot instance segmentation with polar representation,''
  in \emph{CVPR}, 2020.

\bibitem{Ying-CVPR2020}
H.~Ying, Z.~Huang, S.~Liu, T.~Shao, and K.~Zhou, ``Embedmask: Embedding
  coupling for one-stage instance segmentation,'' in \emph{CVPR}, 2020.

\bibitem{Chen-ICCV2019}
X.~Chen, R.~Girshick, K.~He, and P.~Doll{\'a}r, ``Tensormask: A foundation for
  dense object segmentation,'' in \emph{ICCV}, 2019.

\bibitem{Fu-2019}
C.-Y. Fu, M.~Shvets, and A.~C. Berg, ``{RetinaMask}: Learning to predict masks
  improves state-of-the-art single-shot detection for free,'' in \emph{arXiv
  preprint arXiv:1901.03353}, 2019.

\bibitem{Lin2-CVPR2017}
T.-Y. Lin, P.~Goyal, R.~Girshick, K.~He, and P.~Doll{\'a}r, ``Focal loss for
  dense object detection,'' in \emph{ICCV}, 2017, pp. 2980--2988.

\bibitem{Tian-ECCV2020}
Z.~Tian, C.~Shen, and H.~Chen, ``Conditional convolutions for instance
  segmentation,'' in \emph{ECCV}, 2020.

\bibitem{Wang-ECCV2020}
X.~Wang, T.~Kong, C.~Shen, Y.~Jiang, and L.~Li, ``{SOLO}: Segmenting objects by
  locations,'' in \emph{ECCV}, 2020.

\bibitem{Gupta-CVPR2019}
A.~Gupta, P.~Dollar, and R.~Girshick, ``Lvis: A dataset for large vocabulary
  instance segmentation,'' in \emph{CVPR}, June 2019.

\bibitem{Lin-ECCV2014}
T.-Y. Lin, M.~Maire, S.~Belongie, J.~Hays, P.~Perona, D.~Ramanan,
  P.~Doll{\'a}r, and C.~L. Zitnick, ``Microsoft coco: Common objects in
  context,'' in \emph{ECCV}, 2014, pp. 740--755.

\bibitem{Katarzyna-TFML2017}
K.~Janocha and W.~M. Czarnecki, ``On loss functions for deep neural networks in
  classification,'' \emph{Theoretical Foundations of Machine Learning},
  vol.~25, p. 49–59, 2016.

\bibitem{Jurafsky-2009}
D.~Jurafsky, ``Logistic regression,'' in \emph{Speech and Language Processing:
  An Introduction to Natural Language Processing, Computational Linguistics,
  and Speech Recognition}, D.~Jurafsky and J.~H. Martin, Eds.\hskip 1em plus
  0.5em minus 0.4em\relax Prentice Hall, 2009, ch.~5.

\bibitem{Alexey-ICLR2021}
A.~Dosovitskiy, A.~Dosovitskiy, L.~Beyer, A.~Kolesnikov, D.~Weissenborn,
  X.~Zhai, T.~Unterthiner, M.~Dehghani, M.~Minderer, G.~Heigold, S.~Gelly,
  J.~Uszkoreit, and N.~Houlsby, ``An image is worth 16x16 words: Transformers
  for image recognition at scale,'' in \emph{ICLR}, 2021.

\bibitem{He-CVPR2016}
K.~He, X.~Zhang, S.~Ren, and J.~Sun, ``Deep residual learning for image
  recognition,'' in \emph{CVPR}, June 2016, pp. 770--778.

\bibitem{Xie-CVPR2017}
S.~Xie, R.~Girshick, P.~Doll{\'a}r, Z.~Tu, and K.~He, ``Aggregated residual
  transformations for deep neural networks,'' in \emph{CVPR}, 2017, pp.
  1492--1500.

\bibitem{Aixuan-CVPR2021}
A.~Li, J.~Zhang, Y.~Lv, B.~Liu, T.~Zhang, and Y.~Dai, ``Uncertainty-aware joint
  salient object and camouflaged object detection,'' in \emph{CVPR}, 2021.

\end{thebibliography}
\bibliographystyle{IEEEtran}
}

\end{document}